\documentclass[sigconf, nonacm]{acmart}
\usepackage{tikz}
\usetikzlibrary{positioning, arrows.meta}

\usepackage[utf8]{inputenc}
\usepackage{amssymb}
\usepackage{tikz}
\definecolor{white}{rgb}{1,1,1}
\definecolor{black}{rgb}{0,0,0}
\definecolor{grey}{rgb}{0.7,0.7,0.7}
\definecolor{blue}{rgb}{0,0,1}
\definecolor{green}{rgb}{0,1,0}
\definecolor{red}{rgb}{1,0,0}
\definecolor{yellow}{rgb}{1.0, 1.0, 0.0}

\definecolor{lightgrey}{rgb}{0.88,0.88,0.88}
\definecolor{lgrey}{rgb}{0.9,0.9,0.9}
\definecolor{llgrey}{rgb}{0.93,0.93,0.93}
\definecolor{lllgrey}{rgb}{0.96,0.96,0.96}
\definecolor{tableHeadGray}{rgb}{0.85,0.85,0.85}
\definecolor{oddRowGrey}{rgb}{0.95,0.95,0.95}
\definecolor{evenRowGrey}{rgb}{0.85,0.85,0.85}
\definecolor{lightyellow}{rgb}{1.0, 1.0, 0.88}
\definecolor{shadered}{rgb}{1,0.85,0.85}
\definecolor{shadegreen}{rgb}{0.95,1,0.95}
\definecolor{shadeblue}{rgb}{0.95,0.95,1}
\definecolor{selectiveyellow}{rgb}{1.0, 0.73, 0.0}

\definecolor{darkred}{rgb}{0.5,0,0}
\definecolor{darkgrey}{rgb}{0.5,0.5,0.5}
\definecolor{darkgreen}{rgb}{0,0.5,0}
\definecolor{darkblue}{rgb}{0,0,0.5}
\definecolor{darkpurple}{rgb}{0.5,0,0.5}
\definecolor{darkdarkpurple}{rgb}{0.3,0,0.3}

\definecolor{mygreen}{rgb}{0,0.6,0}
\definecolor{myred}{rgb}{0.6,0,0}
\definecolor{mygray}{rgb}{0.5,0.5,0.5}
\definecolor{mymauve}{rgb}{0.58,0,0.82}
\definecolor{myblue}{rgb}{0,0,1}
\definecolor{initblockcolor}{HTML}{F5D590}
\definecolor{decblockcolor}{HTML}{DBE3B5}
\definecolor{propblockcolor}{HTML}{BAD6EC}

\PassOptionsToPackage{colorlinks=false}{hyperref}
\usepackage{hyperref}
\usepackage[acronym]{glossaries-extra}
\glsdisablehyper  
\makeglossaries

\setabbreviationstyle[acronym]{long-short}

\newacronym{ml}{ML}{machine learning}
\newacronym{uq}{UQ}{uncertainty quantification}
\newacronym{dcp}{DCP}{data corruption process}
\newacronym{dc}{data cleaning}{data cleaning}
\newacronym{mv}{MVs}{missing values}
\newglossaryentry{le}{name=label errors,description={}}
\newglossaryentry{sb}{name=selection bias,description={}}
\newacronym{cp}{CP}{conformal prediction}
\newglossaryentry{deb}{name=debiasing,description={}}
\newacronym{bo}{BO}{Bayesian optimization}
\newglossaryentry{bs}{name=beam search,description={}}

\newacronym{DCPT}{DCPT}{data corruption process template}
\newacronym{DCP}{DCP}{data corruption process}

\newacronym{pbf}{XXXXX}{pattern-based corruption mechanism}
\newacronym{pbm}{YYYYY}{pattern-based corruption mechanism}
\newacronym{ep}{\%E}{error percentage}
\newacronym{auc}{AUC}{area under the curve}
\newacronym{mse}{MSE}{mean-squared error}
\newacronym{spd}{SPD}{statistical parity difference}
\newacronym{eo}{EO}{equality of opportunity}
\newacronym{f1}{F1}{F1 score}
\newacronym{jacc}{Jaccard}{Jaccard similarity}

\newacronym{mnar}{MNAR}{missing not at random}
\newacronym{mar}{MAR}{missing at random}
\newacronym{mcar}{MCAR}{missing completely at random}

\usepackage{multirow}
\usepackage{amsfonts}
\usepackage{calc}
\usepackage{paralist}
\usepackage{mdframed}
\usepackage{tikz}
\usetikzlibrary{calc}
\tikzset{
    old inner xsep/.estore in=\oldinnerxsep,
    old inner ysep/.estore in=\oldinnerysep,
    double circle/.style 2 args={
        circle,
        old inner xsep=\pgfkeysvalueof{/pgf/inner xsep},
        old inner ysep=\pgfkeysvalueof{/pgf/inner ysep},
        /pgf/inner xsep=\oldinnerxsep+#1,
        /pgf/inner ysep=\oldinnerysep+#1,
        alias=sourcenode,
        append after command={
        let     \p1 = (sourcenode.center),
                \p2 = (sourcenode.east),
                \n1 = {\x2-\x1-#1-0.5*\pgflinewidth}
        in
            node [inner sep=0pt, draw, circle, minimum width=2*\n1,at=(\p1),#2] {}
        }
    },
    double circle/.default={2pt}{blue}
}

\usepackage{xcolor}
\usepackage{booktabs}
\usetikzlibrary{shapes}
\newcommand{\bubble}[1]{%
  \tikz[baseline=(char.base)]{
    \node[draw, rounded rectangle, fill=white, inner xsep=4pt, inner ysep=2pt] (char) {#1};
  }%
}

\usetikzlibrary{shapes, arrows, positioning}
\usepackage{amsmath}
\usepackage{amssymb}
\usepackage{enumitem}
\usepackage{bm}

\usepackage{graphicx}
\usepackage[linesnumbered,ruled,vlined]{algorithm2e}

\SetCommentSty{mycommfont}

\usepackage{subcaption}
\usepackage{cleveref}
\usepackage{todonotes}

\usetikzlibrary{external}

\usepackage{etoolbox}

\newtoggle{techreport}
\toggletrue{techreport}

\newcommand{\iftechreport}[1]{\iftoggle{techreport}{#1}{}}
\newcommand{\iftechreportelse}[2]{\iftoggle{techreport}{#1}{#2}}
\newcommand{\ifnottechreport}[1]{\iftoggle{techreport}{}{#1}}

\iftechreportelse{
\newcommand{\reva}[1]{{#1}}

 \newcommand{\revc}[1]{{#1}}

\newcommand{\revm}[1]{#1}
\newcommand{\rev}[1]{#1}
}
{
\newcommand{\reva}[1]{{\color{myred}#1}}
\newcommand{\revc}[1]{{\color{mymauve}#1}}
\newcommand{\revm}[1]{{{\color{myblue} {#1}}}}
\newcommand{\rev}[1]{{{\color{myblue} {#1}}}}
}
\newcommand{\argmin}{\mathop{\mathrm{arg\,min}}}
\newcommand{\argmax}{\mathop{\mathrm{arg\,max}}}
\newcommand{\expect}{\mathbb{E}}

\newcommand{\adversemechanism}{\Mechanism^\dagger}
\newcommand{\Dataset}{D} 
\newcommand{\tup}{t}
\newcommand{\ctup}{\cOf{t}}
\newcommand{\CorruptedDataset}{\tilde{D}} 
\newcommand{\Testset}{D_{\text{test}}} 
\newcommand{\Trainset}{D_{\text{train}}} 
\newcommand{\LearningAlgorithm}{\mathcal{A}} 
\newcommand{\Model}{{h}} 

\newcommand{\Performance}{\Psi} 
\newcommand{\Mechanism}{\mathcal{M}} 
\newcommand{\MechanismTemplate}{\boldsymbol{\mathcal{M}}}
\newcommand{\Mfeasible}{\mathbb{M}_{\text{feasible}}} 
\newcommand{\Mspace}{\mathbb{M}}

\newcommand{\dcp}{\text{DCP}\xspace}

\newcommand{\DependencyGraph}{G}
\newcommand{\Vertices}{\mathcal{V}}
\newcommand{\Edges}{\mathcal{E}}
\newcommand{\Noise}{N}
\newcommand{\Noises}{\mathbf{N}}

\newcommand{\CorruptedFeatures}{{\bf A}^*}
\newcommand{\Parameters}{\Theta}
\newcommand{\Bindings}{\theta}
\newcommand{\NoiseDistribution}{\omega}
\newcommand{\NoiseDistributions}{\Omega}
\newcommand{\NoiseDistributionsDom}{\Domain(\Omega)}
\newcommand{\StructuralEquation}{\mathcal{F}}
\newcommand{\StructuralEquations}{\mathbb{F}}
\newcommand{\ConcreteCorruption}{F}
\newcommand{\mb}[1]{\textbf{#1}}
\newcommand{\Parents}[1]{\mb{Pa}(#1)}

\newcommand{\aMV}{\ensuremath{\bot}\xspace}

\newcommand{\Domain}{\mathrm{Dom}}
\newcommand{\Attributes}{\mathbf{A}}
\newcommand{\Attribute}{{A}}

\newcommand{\aval}[2]{{#1}[{#2}]}
\newcommand{\Vset}{\mathbf{x}}
\newcommand{\cOf}[1]{#1^*}

\newcommand{\cAttribute}{{A}^*}

\newcommand{\Targets}{Y}
\newcommand{\cTargets}{\cOf{Y}}

\newcommand{\pattern}{\phi}

\newcommand{\lb}[1]{l_{#1}}
\newcommand{\ub}[1]{u_{#1}}
\newcommand{\PBFdom}[1]{\boldsymbol{\StructuralEquations}[#1]}
\newcommand{\pardom}[1]{\boldsymbol{\Theta}[#1]}
\newcommand{\prob}[0]{p}

\newtheorem{exam}{Example}
\newtheorem{defi}{Definition}
\theoremstyle{definition}

{
  \noindent \textsc{Proof Sketch.}%
}%
{\qedsymbol}

\usepackage{pifont}
\newcommand{\cmark}{\textcolor{darkgreen}{\ding{51}}}
\newcommand{\xmark}{\textcolor{darkred}{\ding{55}}}

\newcommand{\sys}{\textsc{Savage}\xspace}

\newcommand{\gc}{\textsc{GradCancel}\xspace}
\newcommand{\bg}{\textsc{BackGrad}\xspace}
\newcommand{\subpop}{\textsc{SubPop}\xspace}
\newcommand{\rand}{\textsc{Rand}\xspace}

\newcommand{\pr}{{\tt \mathrm{Pr}}}
\newcommand{\SelectionVar}{S}

\newcommand{\dsadult}{Adult\xspace}
\newcommand{\dsemployee}{Employee\xspace}
\newcommand{\dscc}{Credit Card\xspace}
\newcommand{\dsindia}{India Diabetes\xspace}
\newcommand{\dssqf}{SQF\xspace}
\newcommand{\dsdiabetes}{Diabetes\xspace}
\newcommand{\dshmda}{HMDA\xspace}

\newcommand{\sysknnimputer}{KNN-imputer\xspace}
\newcommand{\sysmeanimputer}{mean-imputer\xspace}
\newcommand{\sysmedianimputer}{median-imputer\xspace}
\newcommand{\sysiterativeimputer}{iterative-imputer\xspace}
\newcommand{\sysbc}{BoostClean\xspace}
\newcommand{\sysdiffprep}{Diffprep\xspace}
\newcommand{\syshto}{H2O\xspace}
\newcommand{\sysautosklearn}{AutoSklearn\xspace}
\newcommand{\sysreweighing}{Reweighing\xspace}
\newcommand{\syslfr}{LFR\xspace}
\newcommand{\sysfairsampler}{Fair Sampler\xspace}
\newcommand{\sysfairshift}{Fair Shift\xspace}
\newcommand{\syscpsplit}{Split CP\xspace}
\newcommand{\syscpmda}{CP-MDA-Nested\xspace}
\newcommand{\sysgc}{GradCancel\xspace}
\newcommand{\sysbg}{BackGrad\xspace}

\newcommand{\studyson}{shades-of-null\xspace}

\newcommand{\modellr}{logistic regression\xspace}
\newcommand{\modeldt}{decision trees\xspace}
\newcommand{\Modeldt}{Decision trees\xspace}
\newcommand{\modelrf}{random forest\xspace}
\newcommand{\modelnn}{neural network\xspace}

\newcommand{\mypar}[1]{\smallskip\noindent\textbf{#1}.}

\newenvironment{takeaway}[1]
{\medskip %
  \begin{mdframed}[ %
    linecolor=black, %
    linewidth=1.5pt, %
    roundcorner=10pt, %
    backgroundcolor=black!10 %
    ] %
    \emph{Key Takeaway:}\, #1}
  {\end{mdframed}\par}

\newcommand\vldbdoi{XX.XX/XXX.XX}

\newcommand\vldbavailabilityurl{URL_TO_YOUR_ARTIFACTS}
\newcommand\vldbpagestyle{plain}

\begin{document}
\title{Stress-Testing ML Pipelines with Adversarial Data Corruption}

\settopmatter{authorsperrow=5}
\author{Jiongli Zhu}
\affiliation{\institution{University of California, San Diego} %
 \country{USA}
}
\email{jiz143@ucsd.edu}

\author{Geyang Xu}
\affiliation{\institution{University of California, San Diego} %
  \country{USA}
}
\email{gexu@ucsd.edu}

\author{Felipe Lorenzi}
\affiliation{\institution{University of California, San Diego}
  \country{USA}
}
\email{florenzi@ucsd.edu}

\author{Boris Glavic}
\affiliation{\institution{University of Illinois Chicago} %
  \country{USA}
}
\email{bglavic@uic.edu}

\author{Babak Salimi}
\affiliation{\institution{University of California, San Diego} %
  \country{USA}
}
\email{bsalimi@ucsd.edu}







\begin{abstract}

\revm{
Structured data-quality issues—such as missing values correlated with demographics, culturally biased labels, or systemic selection biases—routinely degrade the reliability of machine-learning pipelines. Regulators now increasingly demand evidence that high-stakes systems can withstand these realistic, interdependent errors, yet current robustness evaluations typically use random or overly simplistic corruptions, leaving worst-case scenarios unexplored.

We introduce \sys{}, a causally inspired framework that (i) formally models realistic data-quality issues through dependency graphs and flexible corruption templates, and (ii) systematically discovers corruption patterns that maximally degrade a target performance metric. \sys{} employs a bi-level optimization approach to efficiently identify vulnerable data subpopulations and fine-tune corruption severity, treating the full ML pipeline, including preprocessing and potentially non-differentiable models, as a black box. Extensive experiments across multiple datasets and ML tasks (data cleaning, fairness-aware learning, uncertainty quantification) demonstrate that even a small fraction (around 5\%) of structured corruptions identified by \sys{} severely impacts model performance, far exceeding random or manually crafted errors, and invalidating core assumptions of existing techniques. Thus, \sys{} provides a practical tool for rigorous pipeline stress-testing, a benchmark for evaluating robustness methods, and actionable guidance for designing more resilient data workflows.}

\end{abstract}


\maketitle

\pagestyle{\vldbpagestyle}


\vspace{-3mm}
\section{Introduction}\label{sec:intro}

\begin{figure}[t]
    \centering
    \includegraphics[width=\linewidth]{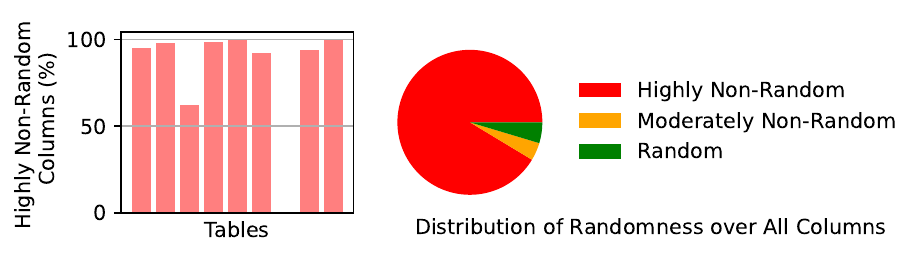}
    \vspace{-.65cm}
    \caption{We analyzed 398 public census tables (>1,100 columns containing gaps). Treating each column's missingness as a binary label and predicting it from other attributes, we found that 91\% of columns achieved an F1 score above 0.9 (bars). This indicates that missing values are predominantly systematic rather than random.
}\vspace{-7mm}
    \label{fig:mv-occurrence}
\end{figure}

\revm{%
Machine-learning pipelines now approve loans, trigger sepsis alerts, and guide parole decisions—roles critical enough that policymakers increasingly demand reliability under "reasonably foreseeable" failures. For instance, Article 15 of the EU Artificial Intelligence Act mandates that high-risk AI systems achieve and maintain appropriate accuracy, robustness, and resilience throughout their lifecycle, while the NIST AI Risk-Management Framework explicitly calls for managing harmful bias and data-quality faults~\cite{EU_AI_Act_2024,NIST_AI_RMF_2024}. Meeting these mandates is challenging because real-world tabular data rarely contain tidy, independent errors. Missing values, label flips, and selection biases typically arise through \emph{structured, interdependent} processes: a "low-risk" flag in medical records suppresses lab tests while correlating with insurance status and demographics~\cite{gianfrancesco2018potential,griffith2020collider}; loan datasets omit repayment histories precisely for subpopulations most likely to default~\cite{plichta2023implications,luo2017validity,ehrhardt2021reject}; and crowdsourced labels drift with cultural nuances, producing systematic misannotations~\cite{haliburton2023investigating}. Figure~\ref{fig:mv-occurrence} confirms how pervasive such structure is—in a large public census corpus, the missingness of values are highly predictable from other attributes. These overlapping errors quietly erode accuracy, fairness, and generalizability~\cite{guhaautomated}, yet without provenance metadata, practitioners have no principled way to certify robustness against such realistic errors.
}

\begin{figure*}
    \centering
    \includegraphics[width=\linewidth]{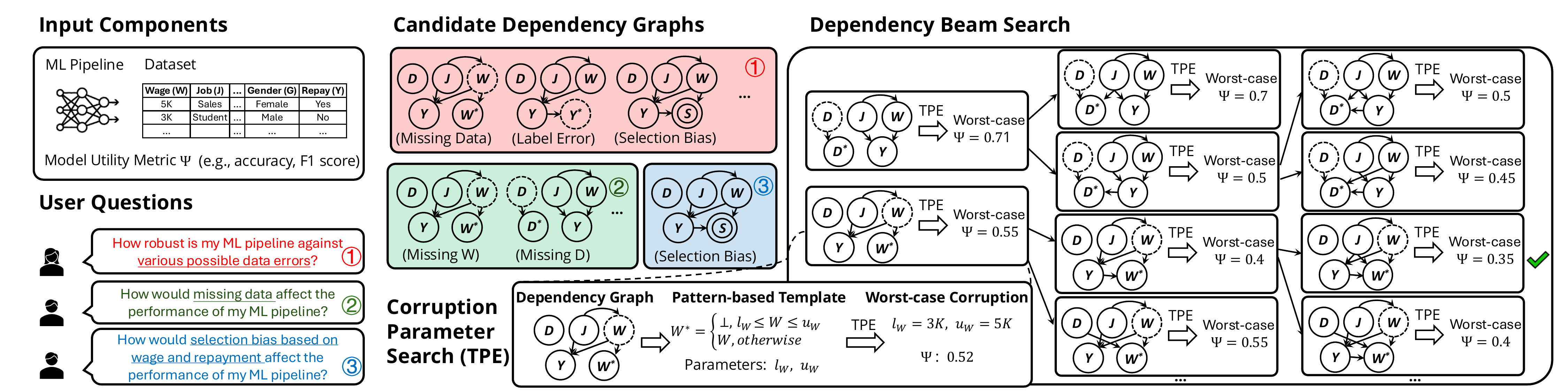}
    \vspace{-6mm}
    \caption{\revm{Overview of \sys. Different user questions map to varying constraints on the dependency graphs, where \textcircled{1} denotes no constraints, \textcircled{2} restricts to missing data, and \textcircled{3} specifies one unique dependency graph for selection bias.
    The dependency beam search is conducted to search the dependency graph and its corresponding worst-case concrete corruption (obtained by TPE) that leads to the lowest model utility measured by $\Psi$.}}
    \vspace{-4mm}
    \label{fig:sys-overview}
\end{figure*}

\revm{To rigorously address this critical issue, this paper introduces \sys (\textbf{S}ensitivity \textbf{A}nalysis \textbf{V}ia \textbf{A}utomatic \textbf{G}eneration of \textbf{E}rrors), a framework that systematically and automatically generates \emph{realistic, high-impact, and adversarial} data corruption scenarios to stress-test end-to-end ML pipelines. Unlike existing benchmarks and adversarial attacks, \sys{} identifies corruption patterns that mirror plausible real-world conditions, for instance, revealing how non-random missingness, label errors, and selection bias jointly exacerbate model failure in vulnerable subpopulations. By automatically discovering these complex yet {\em interpretable} worst-case scenarios, \sys{} helps practitioners and researchers uncover critical pipeline vulnerabilities and develop demonstrably more robust, fair, and trustworthy ML systems.
}

\revm{
Existing robustness benchmarks, including REIN~\cite{abdelaal2023rein}, JENGA~\cite{schelter2021jenga}, CleanML~\cite{li2021cleanml}, and Shades-of-Null~\cite{khan2024still}, typically inject simplistic faults such as uniformly missing values, random label flips, or narrow demographic filters~\cite{guhaautomated,islam2022through}. As summarized in Table~\ref{tab:benchmarks}, these approaches do not systematically explore realistic, structured, and interdependent error patterns, thus underestimating the severity of data-quality issues in high-stakes applications~\cite{li2021cleanml,guhaautomated,abdelaal2023rein}. Indiscriminate data-poisoning attacks also manipulate training data, but they generally target a specific model rather than evaluating entire ML pipelines. Moreover, most poisoning methods, including subpopulation and clean-label attacks~\cite{jagielski2021subpopulation,lu2022indiscriminate,lu2023exploring,munoz2017towards}, are gradient-based, which craft subtle, low-level perturbations that require white-box gradient access, rendering them ineffective when pipelines include non-differentiable preprocessing or must be treated as black boxes.

By contrast, \sys{} produces \emph{explicit, interpretable} corruption patterns—such as missingness or label errors tied to specific demographic or semantic attributes—that mirror interpretable, real-world data-quality issues extensively adopted in ML and data-management studies~\cite{sagadeeva2021sliceline,pradhan2022interpretable,lockhart2021explaining,jagielski2021subpopulation,roy2014formal}.
By treating the entire ML pipeline, including cleaning, feature engineering, and model training, as a black box, \sys{} uncovers worst-case yet plausible failure scenarios that neither existing benchmarks nor model-specific poisoning attacks can reveal.
}

\begin{table}[t]
\centering
\small
\begin{tabular}{l c c c}
\toprule
\textbf{Paper} & \shortstack{\textbf{Data} \\ \textbf{Corruption}} & \shortstack{\textbf{Targeted} \\ \textbf{Errors}} & \shortstack{\textbf{Adversarial} \\ \textbf{Analysis}}\\
\midrule
Budach et al.~\cite{budach2022effects} & \cmark & \xmark & \xmark \\
Islam et al.~\cite{islam2022through} & \cmark & \cmark & \xmark \\
Guha et al.~\cite{guhaautomated} & \xmark & - & - \\
CleanML~\cite{li2021cleanml} & \cmark & \xmark & \xmark \\
REIN~\cite{abdelaal2023rein} & \cmark & \xmark & \xmark \\
JENGA~\cite{schelter2021jenga} & \cmark & \cmark & \xmark \\
Shades-of-Null~\cite{khan2024still} & \cmark & \cmark & \xmark \\
\textbf{\sys (ours)} & \cmark & \cmark & \cmark \\
\bottomrule
\end{tabular}
\caption{Comparison of existing works on data corruption, targeted errors, and adversarial analysis.}
\vspace{-1.2cm}
\label{tab:benchmarks}
\end{table}

\revm{
At the heart of \sys{} is a principled, \emph{causally inspired} framework for modeling realistic data errors that arise from interdependent mechanisms (Section \ref{sec:framework}). These mechanisms are captured in a directed \emph{dependency graph}: nodes represent clean attributes and their corrupted counterparts, while an edge means that the \emph{probability or severity of error} in a child node depends on the \emph{value} of its parent, e.g., if \textsf{insurance-status} is “self-pay,” the \textsf{wage} field is more likely to be recorded as missing. The graph does not encode causal relations among clean attributes; it records only the pathways by which errors propagate. Dependency graphs can be seeded from domain knowledge or discovered automatically with a beam search
(Figure \ref{fig:sys-overview}).
Each graph is paired with flexible, pattern-based \emph{corruption templates} that specify \emph{when} and \emph{how} an attribute is corrupted, such as missingness triggered by demographics or label flips tied to textual cues.
User-defined plausibility constraints (historical frequency thresholds, regulatory rules, validation checks) prune unrealistic scenarios. Together, dependency graphs and corruption templates form an interpretable \emph{Data Corruption Process} that systematically captures common data-quality issues, including missing values, label errors, and selection bias.
}

\revm{
Building on this formal foundation, we formulate the discovery of worst-case corruption scenarios as a \emph{bi-level optimization} problem (Section~\ref{sec:optimization}).
As shown in \Cref{fig:sys-overview}, an upper-level combinatorial beam search~\cite{steinbiss1994improvements} probes the space of all possible error dependencies, e.g., which attributes influence corrupted attributes, that maximally degrades a target metric such as accuracy or fairness.
The lower-level \gls{bo} tunes parameters to maximize performance degradation for a given dependency graph.
Crucially, this approach treats the entire ML pipeline, including preprocessing and training steps, as a black-box function, requiring no gradient information.
To further improve scalability, \sys{} uses a proxy-based strategy: it efficiently identifies harmful corruption patterns using a computationally inexpensive proxy pipeline and transfers these patterns to resource-intensive ML frameworks, achieving substantial runtime improvements (over 10× speed-up on datasets of millions of rows). }

\revm{
To assess \sys{} in practice, we conduct extensive experiments across multiple datasets and ML tasks (Section~\ref{sec:experiments}), thoroughly evaluating the impact of systematically generated corruptions on (i) data-cleaning and preparation methods~\cite{rubin1978multiple,scikit-learn,krishnan2017boostclean,li2023diffprep,ledell2020h2o,feurer2022auto}, (ii) fair and robust learning approaches~\cite{kamiran2012data,zemel2013learning,roh2021sample,roh2023improving}, and (iii) uncertainty quantification techniques~\cite{zaffran2023conformal,lei2018distribution}.
Our results demonstrate that even small, structured corruptions identified by \sys{} can invalidate key assumptions about missingness or label stability, resulting in severe performance degradation that substantially exceeds the impact of random or manually crafted errors.
In summary, our contributions include: (1) a unified, mechanism-aware Data Corruption Process for modeling realistic data-quality errors; (2) a gradient-free, interpretable bi-level optimization method to systematically identify adverse data corruptions; and (3) extensive empirical evidence demonstrating significant, previously unrecognized vulnerabilities in state-of-the-art ML pipelines.}

\section{Related Work}\label{sec:related-work}
\mypar{Benchmarks for Data Quality in \gls{ml}}
As shown in Table~\ref{tab:benchmarks}, most prior work introduces synthetic errors or considers limited corruption scenarios without comprehensive analysis. Guha et al.\cite{guhaautomated} evaluate automated data cleaning but assume identical error distributions in both training and test data, thus failing to capture distribution shifts. JENGA\cite{schelter2021jenga} examines test-time corruptions while keeping training data clean, missing the impact of biased training data. CleanML\cite{li2021cleanml}, REIN\cite{abdelaal2023rein}, and Budach et al.\cite{budach2022effects} study data corruption effects but do not focus on specific error types or detrimental failure modes. Islam et al.\cite{islam2022through} and Shades-of-Null\cite{khan2024still} incorporate targeted corruption mechanisms, yet none perform systematic adversarial analysis to stress-test \gls{ml} pipelines.
\sys fills these gaps by generating structured, realistic data corruptions that reveal failure modes across data cleaning, fair learning, and \gls{uq}. Unlike previous efforts, \sys systematically explores error-generation mechanisms and simulates worst-case corruption scenarios to uncover vulnerabilities that remain hidden under standard benchmark conditions.
By integrating adversarial analysis, \sys enables a more rigorous evaluation of \gls{ml} robustness under structured, non-random corruptions, offering a more realistic assessment of pipeline reliability.

\mypar{Data Poisoning}\revm{
Our work also relates to data poisoning, which deliberately alters training data to degrade model performance, misclassify specific examples, or implant backdoors~\cite{schwarzschild2021just,cina2023wild,jagielski2021subpopulation,lu2022indiscriminate,lu2023exploring,shafahi2018poison,di2022hidden,turner2018clean,chen2017targeted,biggio2012poisoning}.Poisoning methods can be \emph{targeted}, aiming to mislabel particular test instances~\cite{chen2017targeted,turner2018clean}, or \emph{indiscriminate}, broadly reducing overall accuracy~\cite{jagielski2021subpopulation,lu2022indiscriminate}.
Our work shares a high-level goal with indiscriminate data poisoning: both aim to identify corruptions that degrade model performance. However, the motivations, constraints, and techniques differ significantly.
 Poisoning attacks typically seek to evade detection and therefore impose imperceptibility constraints on the perturbations, without requiring the modifications to reflect realistic data quality issues. In contrast, our objective is to evaluate the robustness of ML pipelines with realistic and often systematic data errors.
Unlike poisoning approaches, \sys explicitly models structured corruptions such as selection bias, label errors, and missing values. In addition, our method accommodates both cases where users have limited knowledge of potential data issues and cases where domain-specific error types are known and can be specified.

Another key distinction is that \sys operates on full ML pipelines, including non-differentiable components like imputation or outlier removal. Most poisoning techniques either assume access to a differentiable end-to-end model or ignore preprocessing altogether. While some recent efforts~\cite{liu2021data} incorporate preprocessing (e.g., feature selection), they target narrow scenarios and do not generalize to broader pipeline components.
}

\section{Modeling Data Corruption}
\label{sec:framework}

We now introduce a principled framework for simulating the mechanisms by which real-world data collection processes generate corrupted datasets. Inspired by structural causal modeling~\cite{pearl2009causality,zhu2024overcoming}, this framework explicitly models how systematic dependencies between attributes, noise, and selection processes lead to data-quality issues such as missingness, \gls{le}, and \gls{sb}. Formally, let a dataset \(\Dataset\) consist of \(N\) tuples and have $n$ attributes $\Attributes = \{\Attribute_1, \ldots, \Attribute_n\}$. Given a tuple $\tup$, we use $\aval{t}{\Attribute}$ to denote its value in attribute $\Attribute$.  
For any set of attributes \(\Attributes\), we let \(\Domain(\Attributes)\) denote their joint domain, and \(\Vset\) a specific assignment in that domain. \rev{The domains of all attributes are assumed to contain the special value \aMV that is used to mark  missing values.} The goal of this framework is to model the generation of a corrupted dataset \(\CorruptedDataset\) from \(\Dataset\) by specifying mechanisms that govern how attributes are altered or omitted through noise-driven, interdependent processes. \rev{We use $\Attribute_i$ to denote an attribute in the clean dataset and $\cAttribute_i$ to denote the corresponding corrupted attribute.}

\rev{A \gls{DCP} consists of a \emph{dependency graph} that models at the schema level which attributes of the clean dataset \(\Dataset\) and noise variables modelling stoachstic factors determine the values of the corrupted version $\cAttribute$ of an attribute $\Attribute$. Specifically, for each tuple $\tup$ in the clean dataset, its corrupted counterpart $\ctup$ is created by computing the value of each corrupted attribute $\cAttribute$ based on its parents in the dependency graph using pattern-based corruption functions to be introduced in the following. Noise variables are used to model randomness in the corruption process. For instance, consider a simple label flipping example for a binary label attribute $\Targets$ where the corrupted label $\cTargets$ has a certain probability to being flipped.
That is, the corrupted label is computed solely based on the original label and the value of a noise variable $\Noise$, resulting in the dependency structure: $\Noise \rightarrow \cTargets \leftarrow \Targets$.  As another example, consider that for a subpopulation with specific demographics (attribute $D$), the corrupdated (binary) label is always $0$, and otherwise the corrupted label is equal to the original label:  $D \rightarrow \cTargets \leftarrow Y$. 
}

\begin{defi}[Dependency Graph]
\label{def:depgraph}
A \emph{dependency graph} \(\DependencyGraph = (\Vertices,\Edges)\) is a directed acyclic graph whose vertex set \(\Vertices\) includes:
\begin{itemize}[leftmargin=*]
  \item \emph{Original attributes}, \(\Attributes = \{\Attribute_1,\dots,\Attribute_n\}\),
  \item \emph{Corrupted counterparts}, \(\CorruptedFeatures = \{\cAttribute_1,\dots,\cAttribute_n\}\), where each \(\cAttribute_i\) replaces \(\Attribute_i\) in the corrupted dataset,
  \item \emph{Noise variables}, \(\Noises = \{\Noise_1, \ldots, \Noise_m\}\)
  \item An optional \emph{binary selection indicator}, \(\SelectionVar\), modeling data exclusion.
\end{itemize}
Each directed edge \((u, v) \in \Edges\) specifies that node \(u\) directly influences node \(v\). For a corrupted attribute \(\cAttribute \in \CorruptedFeatures\), its \emph{parent set} \(\Parents{\cAttribute} \subseteq \Vertices\) designates all the variables that govern its corrupted value. We require that (i) each noise variable \(\Noise_i\) is a source in the graph (no incoming edges) and is only connected through outgoing edges to corrupted attributes and (ii) \(\SelectionVar\) and all \(\cAttribute_i\) are sinks.
\end{defi}

\rev{Note that the binary variable \(\SelectionVar\) determines whether a tuple will be included in the corrupted dataset or not. This can be used to model selection bias. 
  For example, consider a medical dataset where patitients with gender attribute $G$ equal to female  have a certain chance (modelled as a noise variable $\Noise_{selbias}$) to be excluded, because of a computer error in the gynecological ward. This corresponds to a graph fragment $\Noise_{selbias} \rightarrow \SelectionVar \leftarrow G$.}
Given a dependency graph \(\DependencyGraph\), which specifies the relationships between the original attributes and how the noise variables and the original attributes affect the corrupted attributes, we now formalize how each corrupted attribute \(\cAttribute \in \CorruptedFeatures\) is computed. 
Rather than defining a single function, \rev{we introduce the concept of a \emph{pattern corruption template}, a parametric family of functions that can model diverse error patterns. Each template is associated with a \emph{selection pattern}, a conjunction of range conditions over the values of the parents of a corrupted attribute. For a given tuple and attribute, the value of the attribute will be corrupted if the pattern evaluates to true for this tuple. The rationale for using templates with parameters is that our system that searches for effective corruptions can be used to find parameter settings for a corruption template that most degrade the performance of a model.}

\begin{defi}[\rev{Pattern} Corruption Function and Template]
\label{def:corruptionfunction}
For an attribute \(\Attribute \in \Attributes\) to be corrupted into $\cAttribute$, a \emph{pattern corruption function} \rev{is a pair $(\ConcreteCorruption, \pattern)$} where:
\[\small
\ConcreteCorruption
\,:\;
\Domain\bigl(\Parents{\cAttribute}\bigr) 
\;\;\longrightarrow\;\;
\Domain(\cAttribute),
\]
where \(\Parents{\cAttribute} \subseteq \Vertices\) is the parent set of \(\cAttribute\) as specified by the dependency graph \(\DependencyGraph\). 
\rev{Furthermore, $\pattern$ is a conjunction of range conditions on the attributes in \(\Parents{\cAttribute}\). For a given tuple $\tup$:
  \[\small
    \pattern(\tup) = \bigwedge_{\Attribute\in\Parents{\cAttribute}}\!
    \Bigl(\lb{\Attribute} \leq \tup(\Attribute) \leq \ub{\Attribute}\Bigr),
  \]
For a clean tuple \(\tup\), the corruption function $(\ConcreteCorruption, \pattern)$ is used to compute the corrupted  value of attribute $\cAttribute$ in the corresponding  corrupted tuple  \(\ctup\):\\[-6mm]
\[\small
  \aval{\ctup}{\cAttribute}
  \;=\;
  \begin{cases}
    \ConcreteCorruption\left(\aval{\tup}{\Parents{\cAttribute}}\right),
                     & \text{if } \pattern(\tup) \\[3pt]
    \aval{\tup}{\Attribute} & \text{otherwise}.
  \end{cases}
\]
}

A \emph{corruption template} \(\StructuralEquations\) for $\cAttribute$ defines a parametric family of corruption functions:
\[
\StructuralEquation:\, \Domain(\Parameters) \longrightarrow \{\ConcreteCorruption,\pattern\},
\]
\rev{where \(\Parameters\) is a set of parameters controlling the behavior of the corruption process.
For all $\Attribute \in \Parents{\cAttribute}$, the bounds    \(\lb{\Attribute}\) and \(\ub{\Attribute}\) are parameters in \(\Parameters\).
By specifying settings $\Bindings$ for $\Parameters$, the template \(\StructuralEquation\) is instantiated into a concrete corruption function \(\StructuralEquation(\Bindings) = (\ConcreteCorruption, \pattern)\).} 
\end{defi}

\rev{Note that the range conditions used in patterns also allow for equality conditions ($\lb{\Attribute} = \ub{\Attribute}$) and one sided comparisons. For convenience we will write such conditions as $\Attribute = c_{\Attribute}$ and $\Attribute \leq c_{\Attribute}$.  
}
Note that while corruption functions are deterministic, randomness in the corruption process is modeled through the noise variables whose values are sampled from a probability distribution.
\rev{Continuing with the selection bias example from above, we model the exclusion of female patients using a pattern $\pattern_\SelectionVar: G = female$ and corruption function $\ConcreteCorruption_\SelectionVar$ with parameter $\prob_{\SelectionVar}$ which determines the probability of exclusion:
  \[\small
    \ConcreteCorruption_{\SelectionVar}(G,\Noise_{selbias}) =
    \begin{cases}
      0 &\text{if}\,\, \Noise_{selbias} \leq \prob_{\SelectionVar}\\
      1 &\text{otherwise}\\
    \end{cases}
  \]
}
\rev{Combining dependency graphs, corruption function templates, and distributions for noise variables we now formally define \glspl{DCPT} and the concrete \glsfmtfullpl{DCP} that result from applying bindings for the parameters of the template. With the exception of values for noise variables, a \gls{DCP} fully specifies the transformation of \(\Dataset\) into \(\CorruptedDataset\) by systematically altering each original attribute based on its specified corruption mechanism and filters tuples based on the value of \(\SelectionVar\).}

\begin{defi}[Data Corruption Process]
\label{def:dcp}
A \emph{\glsfmtlong{DCPT}} \(\MechanismTemplate = (\DependencyGraph,\StructuralEquations,\Parameters)\) is a tuple consisting of:
\begin{itemize}[leftmargin=*]
  \item Dependency graph \(\DependencyGraph\)
  \item Corruption templates $\StructuralEquations$ with a \emph{compatible} template \(\StructuralEquation_{\cAttribute}\) for each corrupted attribute \(\cAttribute \in \CorruptedFeatures\) and a template \(\StructuralEquation_{\SelectionVar}\) for $\SelectionVar$, where compatibility requires that  \(\StructuralEquation_{\cAttribute}\) depends only on \(\Parents{\cAttribute}\), 
  \item A parameter set \(\Parameters\) consisting of $\Parameters_{\cAttribute}$ for each template, used to instantiate \(\StructuralEquation_{\cAttribute}\) into a specific corruption function \(\ConcreteCorruption_{\cAttribute}\). Additionally, \(\Parameters\) contains noise variable distributions \(\NoiseDistributions = \{\NoiseDistribution_1, \ldots, \NoiseDistribution_m\}\), where  $\NoiseDistribution_i$ is the probability distribution for $\Noise_i$.
  \end{itemize}
\rev{  Given bindings \(\Bindings\) for the parameters $\Parameters$ of a \gls{DCPT}  $\MechanismTemplate$, a concrete \emph{\glsfmtlong{DCP}} $\Mechanism = \MechanismTemplate(\Bindings)$ is derived from $\MechanismTemplate$ by applying the bindings to the corruption templates in $\StructuralEquations$ and associating $\Noise_i$ with $\NoiseDistribution_i$.}
\end{defi}

Given a \gls{DCP} \(\Mechanism\), the corrupted dataset \(\CorruptedDataset\) is generated as follows.
For each tuple, values for each noise variables \(\Noise_i\) are first sampled from its distribution \(\NoiseDistribution_i\). The selection function \(\ConcreteCorruption_{\SelectionVar}\) is then evaluated using \(\Parents{\SelectionVar}\) to determine whether the tuple is included in \(\CorruptedDataset\). \rev{If the tuple is included, each corrupted attribute $\cAttribute$ is computed by (i) evaluating $\pattern_{\cAttribute}$ on $\tup$ and if it evaluates to true, apply \(\ConcreteCorruption_{\cAttribute}\) using \(\Parents{\cAttribute}\) to compute the corrupted value. This is done in topological order wrt. the dependency graph \(\DependencyGraph\), ensuring that each attribute is processed only after its parent attributes have been evaluated. Note that two applications of the same \gls{DCP} \(\Mechanism\) may yield different corrupted datasets due to the randomness injected into the process by the noise variables.}
By varying templates, parameters, and noise distributions, \glspl{DCP} model a broad range of data-quality issues, including missingness, \gls{le}, and \gls{sb}, as well as complex multi-attribute dependencies.

We illustrate the application of our framework by modeling two key corruption scenarios: missing data and a compound case involving both label errors and selection bias. These examples demonstrate how corruption processes can be systematically defined using structured dependencies, where missing values arise due to observed and latent factors, and label errors interact with selection mechanisms to shape data availability. The same approach extends naturally to other complex corruption patterns, such as outliers, duplication effects, and interactions between multiple error types.

\begin{figure}[t]
\centering
\begin{subfigure}{\linewidth}
\centering
\begin{minipage}[b]{0.45\linewidth}
\centering
\begin{tikzpicture}[
    every node/.style={draw, circle, minimum size=.7cm, inner sep=0pt},
    cnode/.style={draw, circle, minimum size=.7cm, inner sep=0pt, color=red},
    noise/.style={draw, circle, minimum size=.7cm, inner sep=0pt, color=blue,dashed},
    ->, >=stealth, node distance=1.5cm, thick
]
  \node (Y) [yshift=.9cm, draw] {$Y$};
  \node (YStar) [below of=Y, xshift=-0.75cm,cnode] {$Y^*$};
\node (D) [right of=Y] {$D$};
\node (Ny) [below of=D, noise, xshift=-.75cm] {$\Noise_{Y}$};
  \node (W) [left of=Y, xshift=0.0cm] {$W$};

  \draw[->] (W) -- (Y);
  \draw[->] (D) -- (YStar);
  \draw[->] (Y) -- (YStar);
  \draw[->] (D) -- (Y);
  \draw[->] (Ny) -- (YStar);
\end{tikzpicture}
\end{minipage}%
\hfill
\begin{minipage}[b]{0.45\linewidth}
\centering
\begin{align*}
&D: \text{Demographics}\\
&W: \text{Credit, employment, etc}\\
&Y: \text{Actual repayment}\\
&Y^*: \text{Observed repayment}\\
&\Noise_Y: \text{Noise variable}
\end{align*}
\end{minipage}
\end{subfigure}
\vspace{-3mm}
\caption{Dependency graphs for missing labels.}\label{fig:dep-graph:mv}
\end{figure}
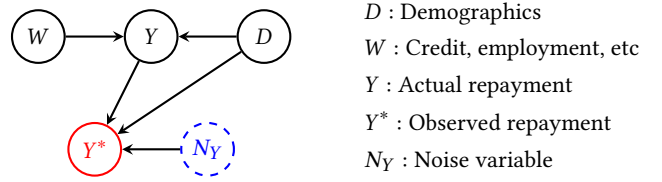

\begin{figure}[t]
\centering
\begin{subfigure}{\linewidth}
  \centering
  \begin{minipage}{0.41\linewidth}
    \centering
\begin{tikzpicture}[
  every node/.style={draw, circle, minimum size=.7cm, inner sep=0pt},
  cnode/.style={draw, circle, minimum size=.7cm, inner sep=0pt, color=red},
  noise/.style={draw, circle, minimum size=.7cm, inner sep=0pt, color=blue,dashed},
  snode/.style={draw, minimum size=.7cm, inner sep=0pt, color=red,double circle={2mm}{red}},
    ->, >=stealth, thick
]
\node (D) at (0,0) {$D$}; 
\node (X) at (1.3,0) {$X$}; 
\node (Z) at (2.6,0) {$Z$}; 
\node (YStar) at (0,0.3) [yshift=.9cm,cnode] {$Y^*$}; 
\node (Y) at (1.3,0.3) [yshift=.9cm, draw] {$Y$}; 
\node (S) at (2.6,0.3) [yshift=.9cm,snode] {$S$}; 
\draw[->] (D) -- (X); 
\draw[->] (D) -- (YStar); 
\draw[->] (X) -- (Y); 
\draw[->] (X) -- (S); 
\draw[->] (Y) -- (YStar); 
\draw[->] (Z) -- (X); 
\draw[->] (Z) -- (S); 
\end{tikzpicture}
\end{minipage} %
\hfill\begin{minipage}{0.49\linewidth}
$D$: Demographics\\
$X$: Socio-cultural traits\\
$Z$: Zipcode\\
$S$: Police patrol indicator\\
$Y$: Actual crime commitment\\
$\cOf{Y}$: Observed crime commitment
  \end{minipage}
\end{subfigure}
\caption{Dependency graphs for label errors and selection bias. For simplicity, we omit the noise variables.}\label{fig:dep-graph:mix}
\end{figure}
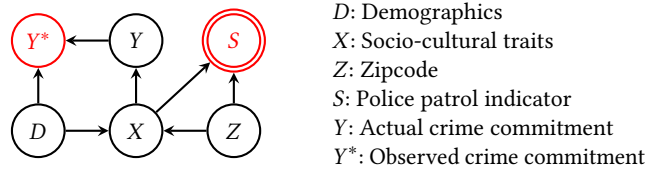


\begin{exam}[Missing Data in Finance]\label{ex:mv-models}
The dependency graph in \Cref{fig:dep-graph:mv} models a \gls{mnar} scenario where missingness in the observed repayment outcome \(Y^*\) depends on the actual repayment status \(Y\) and demographic factors \(D\). \(\Targets\) is corrupted using the following template with parameters \(\{\prob_Y,d, y\}\):
\rev{
\begin{align*}\small
  \ConcreteCorruption_{\cTargets}(\Targets,D,\Noise_Y) &=
                                                         \begin{cases}
                                                           \aMV & \text{if}\,\, \Noise_Y \leq \prob_Y \\
                                                           \Targets &\text{otherwise} \\
                                                         \end{cases}
&\pattern_{\cTargets} &: D = d \land Y = y\\
\end{align*}
}\vspace{-5mm}

where \(\Noise_Y\) is a noise variable taking values in $[0,1]$ with a uniform distribution. The value of \(\Noise_Y\) is compared against the parameter $\prob_Y$ to determine whether $\cOf{Y}$ is missing. For instance, setting \(\prob_Y = 0.95\), \(d = \text{minority}\), and \(y = \text{reject}\) means repayment information is missing with 95\% probability for minorities with rejected applications \rev{and not missing in any other subpopulation.}
\end{exam}

\begin{exam}[Compound Errors in Predictive Policing]\label{ex:mixerr-model}
\Cref{fig:dep-graph:mix} shows a dependency graph for predictive policing, where label errors and selection bias coexist. The actual crime \(Y\) is influenced by socio-cultural traits \(X\) and demographics \(D\), while the observed crime \(Y^*\) is subject to label errors that depend on \(D\) and \(Y\). Selection bias arises from police patrols \(S\), which determine whether data is collected and are influenced by \(X\) and geographic region \(Z\). The label is corrupted using a template with parameters \(\{\prob_Y, d_Y, y_Y\}\):
\rev{
\begin{align*}\small
\ConcreteCorruption_{\cTargets}(\Targets,D) &=
                                              \begin{cases}
                                                1 - \Targets & \text{if}\,\,\Noise_Y \leq \prob_y\\
                                                \Targets &\text{otherwise}\\
                                              \end{cases}
&\pattern_{\cTargets} &:  D = d_Y \land Y = y_Y
\end{align*}
}
where \(\Noise_Y\) is uniformly sampled from $[0,1]$. The parameters \(d_Y\) and \(y_Y\) specify the subpopulation and label values affected.

Selection bias is modelled as a template with parameters \(\{\prob_S, z_S, x_S\}\): %
\rev{ %
  \begin{align*}
  \ConcreteCorruption_{\SelectionVar}(\Noise_{\SelectionVar}) &=
                                                                \begin{cases}
                                                                  0 & \text{if}\,\, \Noise_S  \leq  \prob_S\\
                                                                  1 & \text{otherwise}
                                                                \end{cases}
&\pattern_{\SelectionVar} &: Z = z_S \land X = x_S
  \end{align*}
}
where \(S = 0\) indicates that data was not collected (police did not patrol).
The parameter $\prob_S$ determines the selection probability, while \(z_S\) and \(x_S\) define the region and subpopulation ignored by police.
\end{exam}

\reva{\mypar{Extension to Multi-class Setting}
\Cref{ex:mixerr-model} describes the label error under a binary class setting. 
However, \sys also supports the multiclass label error, which is achieved by segmenting the $\Noise_Y \leq \prob_y$ part into multiple sub-intervals, each representing one class to be changed to.
}

\section{Adversarial Data Corruption}
\label{sec:optimization}

In this section, we develop algorithms for identifying worst-case corruption mechanisms that degrade ML pipeline performance within realistic constraints. Given a set of candidate \glspl{DCPT} \(\Mfeasible\) \rev{provided by the user}, our objective is to determine the most adversarial \gls{DCP} \(\adversemechanism\) that minimizes a performance metric while adhering to structural and domain constraints. \rev{That is, there exists some \gls{DCPT} $\MechanismTemplate  \in \Mfeasible$ and bindings $\Bindings$ for the parameters of $\MechanismTemplate$ such that $\adversemechanism = \MechanismTemplate(\Bindings)$. Abusing notation we will sometimes write $\Mechanism \in \Mfeasible$ to denote that there exists $\MechanismTemplate \in \Mfeasible$ and $\Bindings$ for $\MechanismTemplate$ such that $\MechanismTemplate(\Bindings) = \Mechanism$.}
\rev{As mentioned previously, the user has full control over the specificity of the candidate \glspl{DCP} ranging from letting our approach select the parameters for a fixed corruption function template to searching over a wide range of dependency graphs and candidate corruption templates.}
We define this as an optimization problem over the space of \glspl{DCP} using bi-level optimization to separate template selection from parameter tuning. 


\mypar{Setting}
We are given a training dataset \(\Trainset = \{(x_i, y_i)\}_{i=1}^N\), where \(x_i \in \Domain(\mathbf{X})\) represents the input features of a tuple, and \(y_i \in \Domain(Y)\) denotes the corresponding label. An ML pipeline \(\LearningAlgorithm\) processes \(\Trainset\) through three stages: preprocessing, model training, and post-processing. \rev{As we utilize blackbox optimization techniques, our solution supports arbitrary pipelines.} 
The pipeline produces a  model \(\Model\) based on \(\Trainset\), which is evaluated on a separate test dataset \(\Testset = \{(x_i, y_i)\}_{i=1}^M\). The performance of the model is measured using a task-relevant metric \(\Performance(\Model, \Testset)\), such as accuracy, mean squared error, or fairness measures like demographic parity or equal opportunity. \rev{Without loss of generality we assume that higher values of $\Performance$ indicate better performance.}  To evaluate the effectiveness of a candidate \dcp $\Mechanism$ in degrading $\Performance$, we have to rerun $\LearningAlgorithm$ on $\CorruptedDataset = \Mechanism(\Trainset)$ to get model $\Model_{\CorruptedDataset}$ and reevaluate the metric \(\Performance(\Model_{\CorruptedDataset},\Testset)\).


\mypar{Adversarial Data Corruption}
Given \(\Mfeasible\), 
our objective is to identify the \emph{most adversarial \gls{DCP}} \(\adversemechanism\) that minimizes the  performance metric selected by the user. \rev{Note that for a \gls{DCP} $\Mechanism$, $\Mechanism(\Trainset)$ is not deterministic as we sample values for the noise variables for each tuple. Thus, we optimize for the \gls{DCP} with the lowest expected performance:}

\vspace{-3mm}
\begin{align}\small
    \adversemechanism
    = \argmin_{\Mechanism \in \Mfeasible}
    \expect\left[\Performance\Bigl(\LearningAlgorithm\big (\Mechanism(\Trainset)\big), \Testset\Big)\right].
    \label{eq:optm_problem}
\end{align}\vspace{-3mm}

\mypar{Remark}
The term adversarial in \(\adversemechanism\) does not imply that the identified corruption \glspl{DCP} are rare or extreme. Rather, it refers to a \rev{\glspl{DCP} that  maximally degrades model performance while being within the bounds of what the user considers realistic.
That is, based on the user's background knowledge about possible types of data errors in their domain, we determine the worst case impact on the model performance that can be expected for these types of errors.
The user can provide such background knowledge as input in form of a dependency graph and, potentially also corruption function templates. However, our approach does not require these inputs to be provided, but can also search for dependency graphs and select templates autonomously. By using interpretable patterns, the user can easily judge whether a DCP is realistic and if necessary rerun the system excluding patterns they deem to be unrealistic.
  In contrast to the simple error injection techniques used in past work that evaluates the impact of data quality issues on ML tasks, our approach can ensure the user that their pipeline is robust against realistic worst-case errors.
}

\subsection{Bi-level Optimization Formulation}
\label{sec:optimization:bi-level}

Exploring all candidate \glspl{DCP} -- selecting a dependency graph and corruption template, and bindings for their parameters -- is infeasible. To manage this complexity, we assume a predefined error type (\gls{mv}, \gls{le}, or \gls{sb}) and restrict the search to \glspl{DCPT} for this error type that corrupt a single \emph{target attribute}: $\cAttribute$ for \gls{mv}, $Y^*$ for \gls{le}, or $\SelectionVar$ for \gls{sb}. \rev{Note that the error type determines the corruption function template except for the pattern. Furthermore, we assume that each corrupted attribute $\cAttribute_i$ is associated with a single noise variable $\Noise_i$.} As discussed in \Cref{sec:framework}, a \gls{DCPT} defines a structured corruption mechanism where attributes in the pattern $\pattern$ (used to select the corrupted subpopulation) and used as parameters to the corruption function template correspond to the parents of the target attribute in the dependency graph $\DependencyGraph$, which we assume WLOG includes only these dependencies. \rev{Thus, for a given error type, determining $\StructuralEquation$ except for the pattern, we have to choose the subset of original attributes to be used in the pattern.}  
Despite these restrictions, the number of candidate \glspl{DCPT} remains exponential in the number of attributes as any subset of attributes can be used as the parents of a corrupted attribute, \rev{and for each \gls{DCPT} there may be a large number of possible parameter settings}. To address this, we adopt a \emph{bi-level optimization} framework, where the \emph{upper level} selects the optimal \gls{DCPT}, and the \emph{lower level} tunes parameters, such as the fraction of the selected subpopulation to corrupt. 

\mypar{Structural Components and Corruption Mechanisms}
For a dataset \(\Dataset\) and target attribute $\cAttribute$,
let \(\Mspace = \PBFdom{\Dataset,\cAttribute} \times \pardom{\StructuralEquations} \times \NoiseDistributionsDom\)
is the space of all corruption mechanisms using \glspl{DCP} for the input dataset \(\Dataset\) that above the restrictions mentioned above and their possible parameter settings which include the parameters controlling distributions for all noise variables.

\mypar{Constraints}
Additionally, we allow the specification of further constraints on the candidate space \(\Mspace\). 
Typical constraints include capping the expected fraction of corrupted tuples\footnote{Even though $\Mechanism(\Trainset)$ is probabilistic as noise variables are sampled for each tuple, this expectation is deterministic and can be computed based on the noise distributions and parameters of a corruption template.} via $\expect[\sum_{i=1}^N \mathbf{1}\bigl\{\tup_i \neq \ctup_i \bigr\} \;\le\; k]$,
enforcing valid domain relationships, 
and bounding corruption parameters to plausible intervals. 
We use  \(\Mfeasible \subseteq \Mspace\) to denote the resulting pruned search space.

\mypar{Bi-level Objective}
The optimization objective from \Cref{eq:optm_problem} can be rewritten into
bi-level optimization problem:


\begin{align*}\small
&\min_{\StructuralEquation \in \PBFdom{\Dataset,\cAttribute}}
\;\Biggl\{
\min_{\Parameters \in \pardom{\StructuralEquations} 
  }
\Performance\bigl(\LearningAlgorithm(\Mechanism(\Trainset)), \Testset\bigr)
\Biggr\}\,
  \text{s.t.}\ \Mechanism \in \Mfeasible.
\end{align*}


At the \textbf{upper level}, the pattern $\pattern$ of the \gls{DCPT} is selected (if the error type and, thus, corruption template is fixed). 
The \textbf{lower level}, determines settings $\Bindings$ for the parameters \(\Parameters_{\StructuralEquation}\) (e.g., probabilities or thresholds) and noise distributions \(\NoiseDistributions\) for a given corruption template $\StructuralEquation$ 
to minimize the performance metric \(\Performance\).
\rev{Note that we have dropped the expectation in the bi-level formulation. This is a heuristic choice motivated by the fact that \emph{\gls{bo}} we use for the lower level has been successfully applied in domains where the objective for a solution may be uncertain.}
This bi-level formulation 
balances expressive corruption scenarios with computational feasibility.

\subsection{Solving the Bi-level Optimization Problem}
\label{sec:optimization:solution}

The bi-level optimization process alternates between the \textbf{upper level}, which selects a pattern and the \textbf{lower level}, which tunes parameters for the selected \gls{DCPT} to maximize the degradation of model performance. By iteratively alternating between these two levels, we efficiently navigate the search space while adhering to the constraints defining \(\Mfeasible\).

\mypar{Overall Algorithm}
\Cref{alg:meta-bi-level} provides a high-level overview of the alternating approach. The upper level explores pattern candidates using beam search~\cite{steinbiss1994improvements}, while the lower level applies \glsfmtfull{bo} to refine parameters for each candidate.

\newcommand{\beam}{\mathcal{B}}
\newcommand{\tempcand}{\StructuralEquations_{cand}}
\begin{algorithm}[t]\small
\caption{Alternating Bi-level Optimization}
\label{alg:meta-bi-level}
\KwIn{Training dataset \(\Trainset\), test dataset \(\Testset\), ML pipeline \(\LearningAlgorithm\), feasible parameter space \(\pardom{\StructuralEquations}\), feasible set \(\Mfeasible\), beam width \(B\), number of BO iterations \(\tau\), max beam depth \(d_{\max}\).}
 \(\beam = \emptyset, \tempcand \gets \text{\textsc{determineSeeds}}(\Trainset)\) \tcc*[f]{Init beam}\\
 \For{\(d = 1\) \textbf{\upshape to} \(d_{\max}\)}{
   $\beam_{old} \gets \beam$\\
  \ForEach{\(\StructuralEquation \in \tempcand\)}{
    $\Mechanism \gets \text{\textsc{BO}}(\StructuralEquation)$ \tcc*[f]{ \Cref{alg:bo-param-search} optimizes parameters}\\
    \(\Performance \gets \Performance(\LearningAlgorithm(\Mechanism(\Trainset)), \Testset)\)\\
    $\beam \gets \beam \cup \{(\StructuralEquation, \Mechanism, \Performance)\}$
    }
    $\beam \gets \text{\textsc{top-k}}(\beam, B)$\\
    \If{$\neg\, \text{\upshape\textsc{improves}}(\beam_{old}, \beam)$}{
      $\text{\textsc{break}}$\tcc*[f]{Terminate if no improvement}\\
    }
    $\tempcand \gets \text{\textsc{expand}}(\beam)$ \tcc*[f]{Expand patterns}\\
}
\Return{\(\argmin_{(\StructuralEquation, \Mechanism, \Performance) \in \beam} \Performance\)}
\tcc*[f]{Return optimal $\Mechanism$ from $\beam$}
\end{algorithm}
\subsubsection{Beam Search for Structural Exploration}
\label{sec:beam-search}

To address the exponential size of the search space for \glspl{DCPT}, 
 we employ \emph{beam search}~\cite{steinbiss1994improvements}, a heuristic  search algorithm that balances exploration and exploitation by retaining and expanding only the most promising candidates at each iteration. 
\rev{We maintain two data structures: (i) a set of candidate corruption function templates $\tempcand$ that will be evaluated in the current iteration and a beam $\beam$ that contains triples $(\StructuralEquation, \Mechanism, \Performance)$ where $\StructuralEquation$ is one of the templates we evaluated in the current or previous iterations, $\Mechanism$ is the best \gls{DCP} we have found in the lower level optimization by tuning parameters of $\StructuralEquation$ and $\Performance$ is the performance of the model trained on $\Mechanism(\Trainset)$.
We initialize $\tempcand$ with the set of all single attribute patterns (recall that the corruption function $\ConcreteCorruption_{\StructuralEquation}$ is determined by the error type and we only optimize over the pattern $\pattern_{\StructuralEquation}$ of $\StructuralEquation$). Thus, in our case the beam search is over which attributes to use in the patterns (\rev{Recall that here we assume that each corrupted attribute $\cAttribute_i$ is associated with an independent noise variable $\Noise_i$}).
At each iteration \(d\),  beam search maintains a beam \(\beam\) of size \(B\). Each candidate $\StructuralEquation \in \tempcand$ is evaluated by invoking the lower-level optimization (\Cref{sec:bo}), which tunes parameters including selecting noise distributions \(\Parameters \in \pardom{\StructuralEquations}\) to maximize the degradation of the performance metric \(\Performance\). Once all \glspl{DCPT} in the current candidate set $\tempcand$ have been evaluated and added to the beam, we only retain the top-$B$ performers. The beam's templates after pruning are then  expanded by extending the patterns of each current \glspl{DCPT} in all possible ways with a new attribute. These are the candidate templates for the next iteration.}
%
\iftechreport{

\iftechreportelse{\begin{figure}[t]\small
\centering
\centering
\includegraphics[width=.85\linewidth]{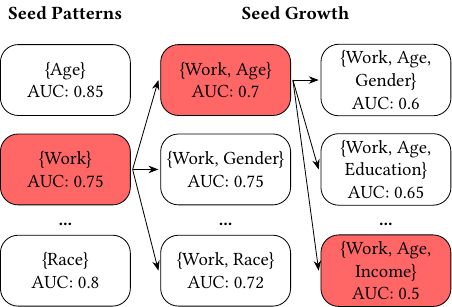}
\caption{Example of beam search (beam size = 1) for finding pattern columns with the lowest AUC of the downstream model.}\label{fig:beam-search}
\vspace{-.5cm}
\end{figure}}{}
\Cref{fig:beam-search} shows an example of the beam search process. We start with single attribute pattern as the seed (e.g., \(\{\text{Age}\}\)), beam search expands patterns with additional attributes.}
\ifnottechreport{As an example, consider the following search.}
Starting with \(\{\text{Work}\}\), the beam may generate candidates \(\{\text{Work, Age}\}\), \(\{\text{Work},\) \(\text{Gender}\}\), and \(\{\text{Work, Race}\}\), each evaluated based on their impact on the performance metric.
Beam search terminates when either no further degradation in \(\Performance\) is observed (\rev{an iteration did not improve the best solution found so far}) or a maximum beam depth \(d_{\max}\) has been reached. By prioritizing the most promising candidates at each step, beam search provides a computationally efficient approach to identifying high-impact structural corruption mechanisms.

\subsubsection{Bayesian Optimization for Parameter Tuning}
\label{sec:bo}

For a fixed \gls{DCP} \(\StructuralEquation\), optimizing parameters \(\Bindings \in \pardom{\StructuralEquation}\) is typically a non-convex optimization problem. Furthermore, it requires evaluating the performance of a parameter setting by running the black-box \gls{ml} pipeline.
\glsfmtfull{bo}~\cite{wang2023recent} is well-suited for this setting, as it balances exploration and exploitation to efficiently locate high-impact parameter configurations \rev{and can be applied in scenarios when the quality of a solution is uncertain to some degree.\footnote{For example, when using \gls{bo} hyperparameter tuning, typically a single model is trained for each hyperparameter setting even though due to randomness in the training process (e.g., random parameter initialization and stochastic gradient descent) the generated optimal for this given setting of the hyperparameters.}}

We employ the \emph{Tree-Structured Parzen Estimator} (TPE)~\cite{bergstra2011algorithms}, a \gls{bo} algorithm that models the parameter space using density estimators.
During initialization, TPE randomly samples a few sets of parameters and estimates their corresponding performance metric \(\Performance\).
At each iteration, TPE separates the set of parameters based on their $\Performance$ values, and fits a probability density function (PDF) for \emph{promising parameters} that result in low \(\Performance\), denoted as $g(\theta)$, and \emph{poor parameters} with high \(\Performance\), denoted as $l(\theta)$.
Then for the next set of parameters to evaluate, TPE chooses the one that maximizes the likelihood ratio:
$\Bindings_t = \argmax_{\Bindings \in \mathcal{P}(\StructuralEquation)} \frac{g(\Bindings)}{l(\Bindings)} = \argmax_{\Bindings \in \pardom{\StructuralEquation}} \frac{\pr(\Bindings \mid \Performance \leq \Performance^*)}{\pr(\Bindings \mid \Performance > \Performance^*)}$,
where \(\Performance^*\) is a quantile threshold of past performance.
\Cref{alg:bo-param-search} shows the full procedure.

\begin{algorithm}[t]\small
\caption{TPE-based Parameter Tuning for Adversarial Mechanisms}
\label{alg:bo-param-search}
\KwIn{Training data \(\Trainset\), test data \(\Testset\), pipeline \(\LearningAlgorithm\), \gls{DCPT} \(\StructuralEquation\), feasible parameter space \(\pardom{\StructuralEquation}\) and \(\NoiseDistributionsDom\), feasible set \(\Mfeasible\), \# iterations \(\tau\).}
\SetAlgoNlRelativeSize{0}
Initialize TPE densities \(g, l\) for parameter space \(\pardom{\StructuralEquation})\)\\
\For{\(t = 1\) \textbf{\upshape to} \(\tau\)}{
    \(\Bindings_t \gets \argmax_{\Bindings} \frac{g(\Bindings)}{l(\Bindings)}\) \tcc*[f]{Sample next parameters}\\
    \(\Mechanism = \StructuralEquation(\Bindings_t)\) \tcc*[f]{Apply parameters}\\
    Project \(\Mechanism\) to \(\Mfeasible\) \\
    \(\Performance_t = \Performance(\LearningAlgorithm(\Mechanism(\Trainset)), \Testset)\) \tcc*[f]{Evaluate parameters}\\
    Update TPE densities \(g, l\) based on \(\Performance_t\)\\
}
\Return{\(\argmin_{t} \Performance_{t}\)}
\end{algorithm}


\subsubsection{Implementation and Efficiency Enhancements}
\label{sec:implementation-enhancements}

Although the bi-level approach finds adversarial corruption mechanisms effectively, we integrate the following strategies to further enhance efficiency and scalability:

\mypar{Heuristics}
During beam search, we impose domain-informed heuristics to avoid unproductive expansions. For instance, pattern-based templates must always include the target attribute(s) (e.g., in Missing-Not-At-Random settings) and the label attribute, which causes a compound of covariate shift and concept drift. We also limit the number of attributes used in patterns to prevent overly sparse subpopulations and enforce feasibility rules that reflect domain constraints (e.g., compatible attribute interactions). By restricting the structural search in this manner, we prune large portions of the search space while preserving high-impact corruption mechanisms.

\iftechreport{
\mypar{Parallel Evaluations}
Both the upper-level beam search and the lower-level \gls{bo} are amenable to parallelization. Candidate structures in the beam can be expanded and evaluated concurrently, and parameter configurations for each candidate can also be tested in parallel. This parallel design significantly reduces runtime on distributed architectures, since multiple structural or parametric evaluations are conducted simultaneously.
}

\mypar{Knowledge Reuse and Warm-Starting}
To reduce computational overhead, we reuse structural and parametric insights gleaned from simpler pipelines or smaller data samples. This reuse, or ``warm-starting'', leverages the observation that many core properties of adversarial corruption mechanisms remain applicable across different dataset scales and pipelines. For instance, dependency graphs and corruption templates identified with a lightweight model can serve as valuable initial structural candidates when transitioning to a more computationally intensive pipeline.
Likewise, parameter distributions (e.g., from TPE density estimators) learned on smaller data can provide an effective initialization for \gls{bo} on larger data, thereby expediting convergence.




\section{Experiments}\label{sec:experiments}

In the experiments, we answer the following research questions:
%
\textbf{Q1:} How do data errors affect the accuracy and fairness of models trained on cleaned datasets prepared with state-of-the-art data cleaning algorithms? Furthermore, are methods sensitive to particular data corruption processes and types of errors
 (\Cref{sec:exp:impact})?
\textbf{Q2:} Can robust learning algorithms produce robust models over dirty data and which characteristics of the data corruption process affect their success? Do the guarantees of \glsfmtfull{uq} methods still hold when the data is subject to systematic errors (\Cref{sec:benchmark:robust-fair})?
\textbf{Q3:} How robust are models when data corruption is adversarial and systematic compared to non-adversarial settings as used in prior experimental studies on the impact of data quality and cleaning on model robustness~\cite{khan2024still}  (\Cref{sec:exp:case-study})? 
 \textbf{Q4:} What is the effectiveness and performance of \sys and its components? \revm{And how effective is \sys compared to state-of-the-art data poisoning techniques?} (\Cref{sec:effic-effect-sys}).
%
All our experiments are performed on a machine with an AMD Opteron(tm) 4238 processor, 16 cores, and 125G RAM.
Experiments are repeated 5 times with different random seeds, and we report the mean (error bars denote standard deviation).
The source code for \sys is available at \url{https://github.com/lodino/savage}.

\vspace{-2mm}
\subsection{Setup}\label{sec:experiment-settings}
\vspace{-1mm}
\mypar{Datasets and Data Errors}
As shown in \Cref{tab:datasets}, we conduct experiments primarily on six representative \gls{ml} tasks: classification (\emph{\dsadult, \dsemployee, \dscc, \dsindia}, and \emph{SQF} datasets) used for the evaluation of \gls{dc} and robust learning algorithms, and one regression dataset (\emph{\dsdiabetes}) used for evaluating \gls{uq} methods.
We use the \emph{\dsindia} dataset for a case study on error patterns as it has been analyzed in related work~\cite{khan2024still}.
Although \sys supports corruption with a wide range of errors, in the paper, we focus on three common data errors: \gls{mv}, \gls{sb} (and sampling error), and \gls{le}.
Unless explicitly mentioned, the target column for injecting \gls{mv} is automatically selected during beam search and we attack only the training data.

\begin{table}[t]\small \centering
\begin{tabular}{@{}lcccc@{}}
\toprule
    \textbf{Dataset} & \# rows & \# cols & Label           & Task           \\
    \toprule
    \dsadult         & 45K     & 3       & Income>\$50K    & Classification \\
    \dsemployee      & 4.7K    & 9       & Resignation     & Classification \\
    \dscc            & 30K     & 8       & Default         & Classification \\
    \dsindia         & 905     & 17      & Type 2 Diabetes & Classification \\
    SQF              & 48K     & 14      & Frisk           & Classification \\
    \revc{\dshmda}      & \revc{3.2M}     & \revc{8}      & \revc{Loan Approval}        & \revc{Classification}     \\
    \dsdiabetes      & 442     & 10      & Severity        & Regression     \\
\bottomrule
\end{tabular}
\caption{Datasets and \gls{ml} tasks.}
\vspace{-7mm}
\label{tab:datasets}
\end{table}

\begin{table}[t]\small
\centering
\begin{tabular}{@{}ll@{}}
\toprule
\textbf{Algorithm}                           & \textbf{Targeted Error Types} \\ \midrule
\text{Imputers}~\cite{scikit-learn}          & missing values                 \\
\text{\sysbc}~\cite{krishnan2017boostclean}  & missing values , selection bias, label errors                        \\
\text{\sysdiffprep}~\cite{li2023diffprep}    & missing values, outliers                         \\
\text{\syshto}~\cite{cook2016practical}      & missing values                         \\
\text{\sysautosklearn}~\cite{feurer2022auto} & missing values, selection bias                         \\
\bottomrule
\end{tabular}
\caption{\Glsentrylong{dc} algorithms.}
\label{tab:exp_cleaners}
\vspace{-8mm}
\end{table}

\begin{table}[t]\small
\centering
\begin{tabular}{@{}llc@{}}
\toprule
\textbf{Algorithm}                              & \textbf{Objective} & \textbf{Targeted Error Types} \\ \midrule
\text{\sysreweighing}~\cite{kamiran}            & \text{Debiasing}   & \xmark                         \\
\text{\syslfr}~\cite{zemel2013learning}         & \text{Debiasing}   & \xmark                         \\
\text{\sysfairsampler}~\cite{roh2021sample}     & \text{Debiasing}   & label errors                         \\
\text{\sysfairshift}~\cite{roh2023improving}    & \text{Debiasing}   & correlation shift                         \\
\hline
\text{\syscpsplit}~\cite{lei2018distribution}   & \text{\gls{uq}}    & \xmark                         \\
\text{Split-MDA CP}~\cite{zaffran2023conformal} & \text{\gls{uq}}    & missing values                         \\
\bottomrule
\end{tabular}
\caption{Debiasing and \gls{uq} algorithms.}
\label{tab:exp_methods}
\vspace{-10mm}
\end{table}
\mypar{Algorithms and Models}
\Cref{tab:exp_cleaners} lists the data cleaning solutions used in our experiments. For each approach, we specify which types of errors are targeted by the method. To cover a wide range of methods, we included automated systems like \sysbc~\cite{krishnan2017boostclean}, \syshto~\cite{ledell2020h2o}, \sysdiffprep~\cite{li2023diffprep}~\footnote{The \sysdiffprep source code only contains \modellr models. Also, it does not support non-differentiable models such as \modeldt and \modelrf. Therefore, we only test it with \modellr as downstream models.}, and \sysautosklearn~\cite{feurer2022auto}.
We also use popular implementations of standard imputation techniques, including imputing with the mean and median value of a feature, and advanced methods such as KNN imputation and iterative imputation~\cite{scikit-learn}.

We also evaluate the robustness of techniques that aim to reduce biases of models or quantify the uncertainty in model predictions.
To evaluate how well debiasing and \glsfmtfull{uq} algorithms handle data errors, we contrast approaches that were explicitly designed to handle data errors with those that do not.
We evaluate the techniques listed in \Cref{tab:exp_methods}.
The purpose of debiasing techniques is to reduce biases in predictions made by a model.
Specifically, for the debiasing tasks, we test two widely used preprocessing methods: \sysreweighing ~\cite{kamiran} and \syslfr~\cite{zemel2013learning}, as well as \sysfairsampler~\cite{roh2021sample}, which addresses noisy labels, and \sysfairshift~\cite{roh2023improving}, which handles correlation shifts where the correlation between the label and sensitive attribute changes. For \gls{uq}, we use two \gls{cp} techniques: \syscpsplit~\cite{lei2018distribution} and \syscpmda~\cite{zaffran2023conformal}. For regression, \gls{cp} takes as input a significance level $\alpha$ in $[0,1]$ and returns, for each data point, a prediction interval such that, with $1 - \alpha$ probability, the data point's true label is within the interval.

We evaluate how these cleaning, debiasing, and \gls{uq} algorithms are impacted by systematic \gls{dcp} by training \gls{ml} models on data prepared using these methods. We consider the following types of models: \emph{\modellr}, \emph{\modeldt}, \emph{\modelrf}, and \emph{\modelnn}, which is a feed-forward neural network with 1 hidden layer containing 10 neurons.
The parameters of models are specified in the code repository~\cite{savagecode}.

\revm{
As discussed in~\cref{sec:related-work},
  \sys and indiscriminate data poisoning both attack a model's performance by corrupting the training data.
Even though the motivations (and requirements) of these two lines of work are different, to evaluate the raw effectiveness in degrading model performance, we compare \sys against the state-of-the-art Gradient Cancelling~\cite{lu2023exploring} (\gc) and Back Gradient~\cite{munoz2017towards} (\bg) poisoning attacks.
\iftechreport{Another relevant existing work is the subpopulation attack~\cite{jagielski2021subpopulation} (\subpop), which investigates the effect of corruption within subpopulations.
Different from our primary setting, they aim to reduce model utility for some subpopulation, by appending mislabeled points to that subpopulation.
}
}

\mypar{Metrics}
We measure the amount of errors in a dataset as the percentage \%E of the rows that are affected by at least one errors. For instance, for missing values, 50\% would indicate that 50\% of the rows contain one or more missing values.
To measure model performance, we use \gls{auc} \revm{and \gls{f1}} for classification tasks and \gls{mse} for regression tasks.
We also measure the bias of a model using standard fairness metrics:  \gls{spd}~\cite{dwork2012fairness} and \gls{eo}~\cite{hardt2016equality}.
For \gls{uq} tasks, we calculate coverage rate.

\mypar{Dependency Graph Transfer}
To deal with the high runtime for automated data-cleaning frameworks such as AutoSklearn and BoostClean, we warm start the search for \sysbc, \sysautosklearn, \sysdiffprep, and \syshto by transferring the worst-case dependency from the \sysiterativeimputer  and then finetune the corruption parameters using TPE. 
\begin{figure*}[t]
    \centering
    \includegraphics[width=\linewidth]{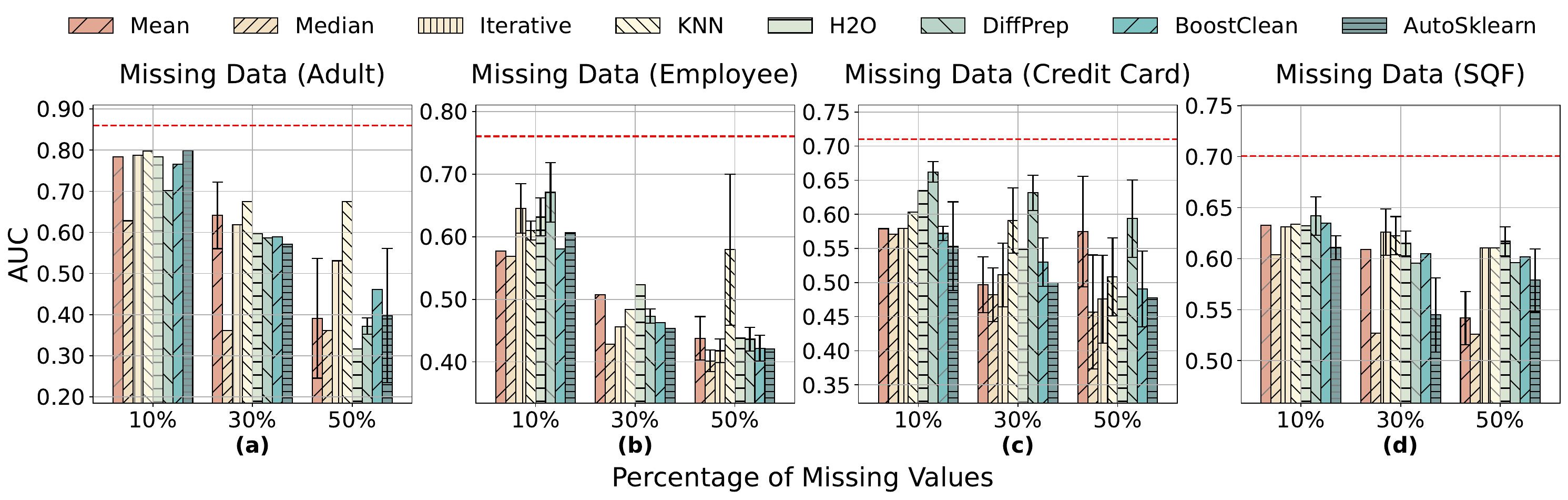}
    \vspace{-8mm}
    \caption{AUC of \modellr when corrupting \dsadult (a), \dsemployee (b), \dscc (c) and SQF (d) datasets with \gls{mv}.}
    \label{fig:benchmark:mv-attack}
    \vspace{-3mm}
\end{figure*}



\begin{figure*}[t]
    \centering
    \includegraphics[width=\linewidth]{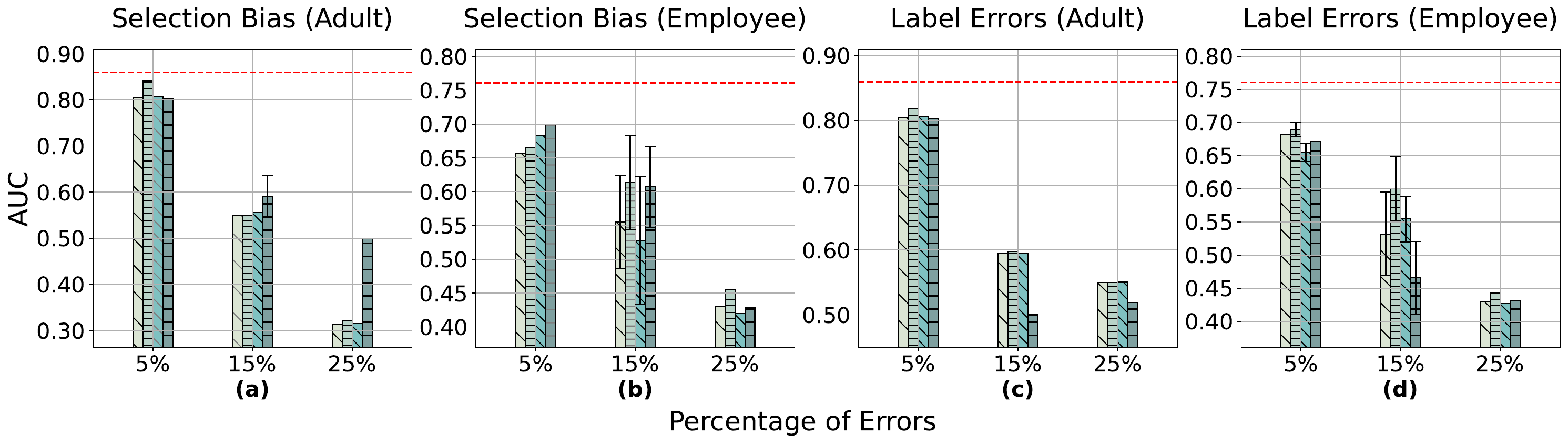}
    \vspace{-8mm}
    \caption{AUC of \modellr, corrupting \dsadult (a, c) and \dsemployee (b, d)  with \gls{sb} (left) and \gls{le} (right).}
    \label{fig:benchmark:sb-le-attack}
    \vspace{-5mm}
\end{figure*}


\begin{figure}[t]
    \centering
    \includegraphics[width=\linewidth]{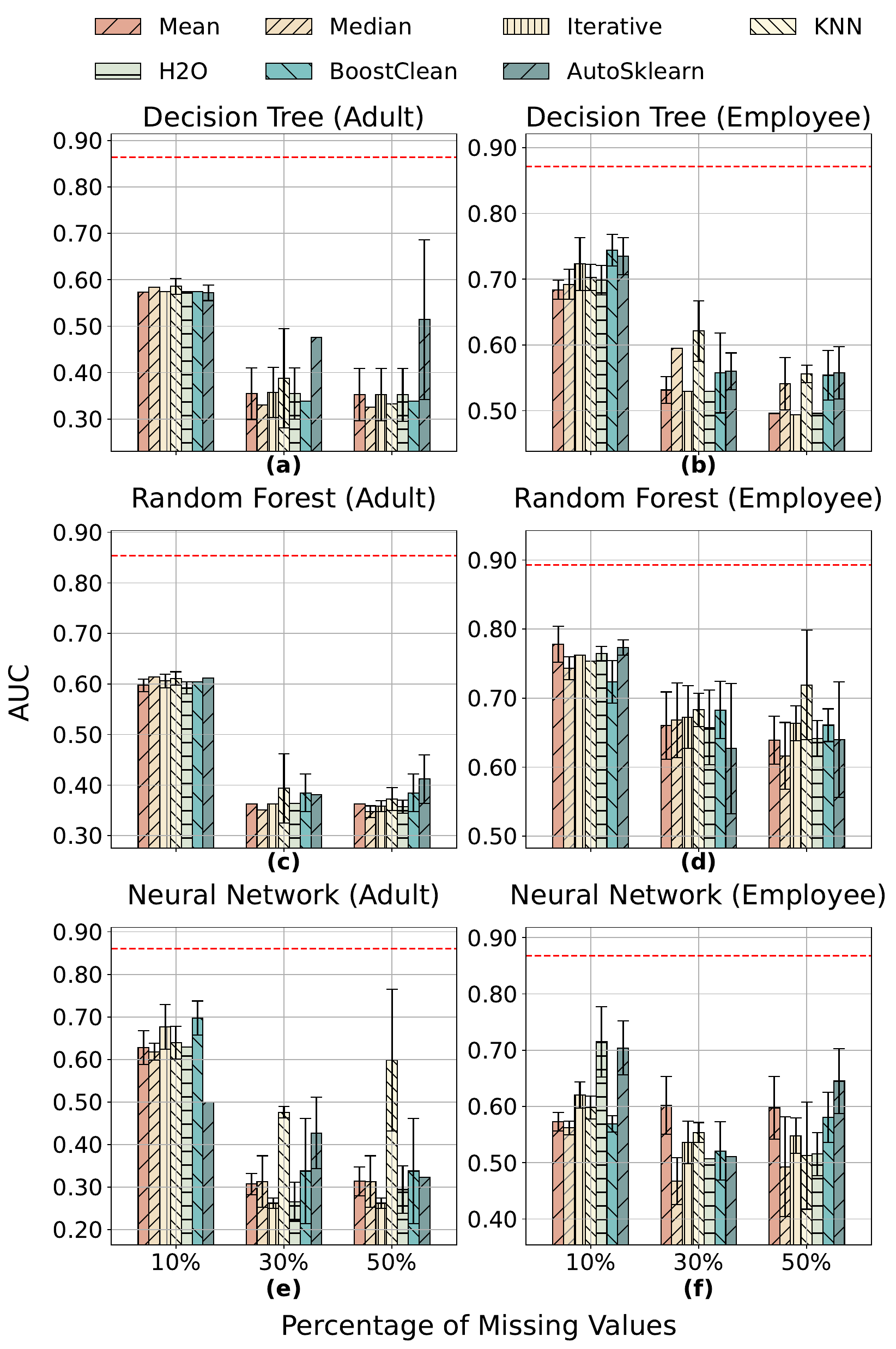}
    \vspace{-8mm}
    \caption{\gls{auc} of \modeldt (first row), \modelrf (second row), and \modelnn (third row) when injecting \gls{mv} into the \dsadult (left) and \dsemployee (right).}
    \label{fig:benchmark:downstream-adult-employee}
    \vspace{-6mm}
\end{figure}

\subsection{Sensitivity of Data Cleaning Methods}\label{sec:exp:impact}
We use \sys to inject errors into datasets to attack automated data cleaning techniques (\Cref{sec:benchmark:data-cleaning}), robust fairness algorithms (\Cref{sec:benchmark:robust-fair}), and \gls{uq} (\Cref{sec:benchmark:uq}). 
We vary \gls{ep} and measure 
model accuracy and fairness.

\subsubsection{Data Cleaning Techniques}\label{sec:benchmark:data-cleaning}

The results for data cleaning techniques are shown in \Cref{fig:benchmark:mv-attack,fig:benchmark:sb-le-attack,fig:benchmark:downstream-adult-employee}, where the red dotted line represents the maximum \gls{auc} achieved across all the methods when no errors are present in the data. We use this as a baseline as some of the techniques apply transformations that are beneficial even if no data errors are present.

\mypar{Varying Data Quality Issues}
We first focus on \modellr, varying the type of data quality issues injected by \sys  using the \dsadult, \dscc, \dssqf datasets.
\iftechreportelse{
For \gls{mv}  (\Cref{fig:benchmark:mv-attack}), less than 10\% \gls{mv} in a single column is sufficient to cause a drop of more than 0.05 in \gls{auc} for most methods, except for \sysdiffprep on the \dscc dataset.
At 30\% missing values, \gls{auc} drops exceed 0.15 across most scenarios except for \sysdiffprep and \sysknnimputer on the \dscc dataset, and methods on the \dssqf dataset, which contains less highly-predictive features for data errors to target and harm \gls{auc} significantly.
Advanced methods, including those which learn a model to impute values such as \sysknnimputer and \sysiterativeimputer, or model-dependent cleaning methods like \sysdiffprep, tend to exhibit greater robustness compared to simpler approaches such \sysmeanimputer that do not adjust cleaning to achieve a particular target.
For example, \sysdiffprep achieves the best \gls{auc} on the \dscc dataset (\Cref{fig:benchmark:mv-attack}c), while \sysknnimputer performs exceptionally well on the \dsadult dataset (\Cref{fig:benchmark:mv-attack}a).
\sysmedianimputer performs worse then \sysmeanimputer in all experiments. While this may seem counterintuitive, given that the median is less sensitive to outliers than the mean, for binary features, we can often shift the median from $1$ to $0$ or vice versa while the mean can only be impacted by at most \gls{ep}.

\Gls{sb} and \gls{le} have a more detrimental effect on model performance than \gls{mv} for the same error budget. For this experiment, we excluded the missing value imputation techniques.
On the \dsemployee dataset, with a 25\% error budget, both error types cause the \gls{auc} of all cleaning techniques to drop below 0.45, which is comparable to the impact of 50\% \gls{mv}.
Similarly, on the \dsadult dataset, a 15\% error budget reduces the \gls{auc} of all methods to below 0.6, which is more severe than the effect of 30\% \gls{mv}.
This difference in adverse effects arises because \gls{sb} and \gls{le} introduce larger shifts in $\pr({\text{Label}\mid\text{Covariate}})$, and disrupt the relationships between features and labels more significantly than \gls{mv}.

For small amounts of \gls{sb} and \gls{le}, \sysdiffprep exhibits the best overall performance.
For instance, when the error budget is 5\%, it has the highest AUC in 3 out of 4 cases.
In contrast, \sysautosklearn is sensitive to this type of error, performing well under selection bias but performing poorly in other scenarios.
\syshto, while never achieving the best performance, remains relatively stable across all conditions. This stability is due to its straightforward data processing approach, which simply combines mean imputation and standardization, thus reducing overfitting to spurious correlations caused by data quality issues.
}
{For \gls{mv} (\Cref{fig:benchmark:mv-attack}), fewer than 10\% missing values in a single column can reduce \gls{auc} by over 0.05 for most methods, with the exception of \sysdiffprep on \dscc. At 30\%, the reduction exceeds 0.15 in most cases, except for \sysdiffprep and \sysknnimputer on \dscc, and all methods on \dssqf, which contains fewer predictive features.
Learning-based methods like \sysknnimputer and \sysdiffprep are generally more robust than simpler techniques such as \sysmeanimputer.
\iftechreport{For instance, \sysdiffprep achieves the highest \gls{auc} on \dscc, while \sysknnimputer performs best on \dsadult.}
Compared to \gls{mv}, \gls{sb} and \gls{le} are more detrimental under the same budget.
For this experiment, we exclude imputation methods.
On \dsemployee, a 25\% corruption budget reduces all methods below 0.45 \gls{auc}, comparable to the impact of 50\% \gls{mv}. Similarly, on \dsadult, 15\% corruption leads to \gls{auc} below 0.6. These errors are more harmful as they induce greater shifts in $\pr(\text{Label}\mid\text{Covariate})$, disrupting feature-label dependencies more severely.
}

\begin{takeaway}
  Data-cleaning techniques are sensitive to small amounts of systematic data corruption. Adaptive techniques like \sysdiffprep are more effective, but also less stable.
\end{takeaway}

\mypar{Varying Downstream Models}
Next, we vary what type of model that is trained using the \dsadult and \dsemployee datasets (\Cref{fig:benchmark:downstream-adult-employee}).
\iftechreportelse{
In general, \modeldt and \modelnn models exhibit higher variance.
For example, on the \dsadult dataset with a 30\% error budget, the difference between the best and worst \gls{auc} values is 0.2 for \modeldt and 0.3 for \modelnn, while it is only 0.1 for \modelrf.
This indicates that \modeldt and \modelnn models are more sensitive to varying transformations of the data, making them more prone to overfitting.
The \modelrf models, in contrast, demonstrate reduced variance and greater robustness due to their ensemble nature, which mitigates overfitting.
This benefit is particularly significant on the \dsemployee dataset (Figure~\ref{fig:benchmark:downstream-adult-employee}d), where compared to \modeldt and \modelnn, methods coupled with \modelrf consistently maintain more than 0.1 \gls{auc} values across different levels of data corruption.

Interestingly, \modeldt and \modelrf models show similar rankings of data cleaning techniques across both datasets.
This similarity arises from the shared internal mechanisms of tree-based models, such as recursive partitioning and split-based decision rules. This trend is particularly clear on the \dsadult dataset, where the top-performing methods for \modeldt and \modelrf are consistent across all error budgets.

\begin{takeaway}
\Modeldt and \modelnn models are more sensitive to data quality issues, exhibiting higher variance in performance, while \modelrf benefits from reduced overfitting and demonstrates greater robustness.
Similar downstream models share a similar response to the same data cleaning techniques.
\end{takeaway}
}
{
In general, \modeldt and \modelnn exhibit greater variance in performance. For example, on \dsadult with 30\% corruption, the \gls{auc} range spans 0.2 for \modeldt and 0.3 for \modelnn, compared to only 0.1 for \modelrf. This indicates greater sensitivity and potential overfitting in \modeldt and \modelnn.
By contrast, \modelrf demonstrates higher robustness, likely due to its ensemble structure, with consistently higher \gls{auc} across corruption levels—particularly on \dsemployee.
\iftechreport{Interestingly, \modeldt and \modelrf produce similar rankings of cleaning techniques, likely due to shared tree-based mechanisms such as recursive partitioning and split-based decision rules. This pattern is especially evident on \dsadult.}

\begin{takeaway}
\Modeldt and \modelnn are more susceptible to data corruption than \modelrf. 
\iftechreport{Architecturally similar models tend to respond similarly to data cleaning strategies.}
\end{takeaway}
}

\subsection{Debiasing and Uncertainty Quantification}\label{sec:benchmark:robust-fair}
We analyze the robustness of fair \gls{ml} and \gls{uq} under systematic \gls{le} and \gls{sb}.

\subsubsection{Debiasing}
\label{sec:debiasing}

\iftechreportelse
{
For this experiment, we use the \dsadult dataset with \modellr as the downstream model.
We evaluate \sysfairsampler~\cite{roh2021sample}, which is designed to handle \gls{le}, and \sysfairshift~\cite{roh2023improving}, which addresses correlation shifts.
To simulate their application scenarios, we use \sys to (1) generate systematic label flips for evaluating \sysfairsampler, and (2) introduce \gls{sb} to test \sysfairshift.
In addition, for comparison will also include \sysreweighing and \emph{Learning Fair Representations} (\syslfr) which are not designed to handle data corruption.
We evaluate each method based on the \gls{f1} and unfairness, measured by \gls{eo}, after introducing data corruptions with a 10\% budget.
For reference, we also report the \gls{f1} and \gls{eo} of a \modellr classifier trained directly on the corrupted data (\emph{Orig.}). Without data corruption, the \gls{f1} of a regular \modellr is 0.555, and the \sysfairsampler, \sysfairshift, \sysreweighing, \syslfr exhibit an \gls{f1} of at least 0.471.

The results are shown in~\Cref{tab:robustness-label-bias}. Training a model on the corrupted data directly already has a low \gls{f1} of 0.32.
In general, \gls{le} with a 10\% budget significantly degrades both accuracy and fairness across all models. Specifically, none of the approaches achieves an \gls{f1} of above 0.21.
\sysreweighing fails to ensure fairness, with a highly exacerbated \gls{eo} of 0.49, indicating its poor robustness to \gls{le}.
While \syslfr achieves the lowest \gls{eo} of 0.01, the resulting model's performance is very poor: an \gls{f1} of 0.01.
\sysfairsampler achieved decent \gls{eo} at an \gls{f1} of 0.20 that is close to that of \sysreweighing. Thus, while \sysfairsampler performs better than the alternatives that have not been designed to deal with \gls{le}, the resulting model's \gls{f1} is still very poor. This demonstrates that even techniques designed to deal with particular data errors often fail if the errors are systematic.

The results for \gls{sb} are shown in~\Cref{tab:robustness-selection-bias}. Similar to \gls{le}, \gls{sb} adversely impacts both accuracy and fairness, though the severity of the impact varies across methods. \sysfairshift demonstrates the best balance between accuracy and fairness, achieving an \gls{eo} of 0.13 and an \gls{f1} of 0.35.
However, \sysfairshift does not guarantee perfect fairness as it assumes the marginal probabilities of labels and sensitive attributes remain unchanged -- an assumption that is often violated in real-world settings and is mildly broken in these experiments.
\sysreweighing actually increases the bias of the model at an \gls{eo} of 0.36, indicating limited robustness to unfairness caused by \gls{sb}, though its \gls{f1} of 0.33 is slightly better.
\syslfr again fails to produce a usable classifier due to its near-zero \gls{f1} of 0.01.

\begin{takeaway}
  While advanced debiasing techniques such as \sysfairsampler and \sysfairshift that are designed to handle certain types of errors are effective at reducing unfairness in the presence of data corruption, they do not address reduced classification performance (\gls{f1}) when errors are systematic, leading to models that perform too poorly to be of any practical relevance.
\end{takeaway}
}
{
We evaluate debiasing methods on the \dsadult dataset using \modellr. Specifically, we test \sysfairsampler~\cite{roh2021sample}, which addresses \gls{le}, and \sysfairshift~\cite{roh2023improving}, designed for correlation shifts. To simulate their application scenarios, we use \sys to (1) generate systematic label flips for evaluating \sysfairsampler and (2) introduce \gls{sb} for assessing \sysfairshift. For comparison, we also include \sysreweighing and \syslfr, which are not designed for systematic corruption. We use a 10\% corruption budget using \gls{f1} and unfairness measured by \gls{eo}. We also report performance of a baseline \modellr trained on the corrupted data (denoted \emph{Orig.}). Without corruption, all methods achieve \gls{f1} $\geq$ 0.471, with regular \modellr reaching 0.555.

\Cref{tab:robustness-label-bias} shows the results under \gls{le}. Training directly on corrupted data yields a low \gls{f1} of 0.32. All methods perform poorly under \gls{le}, with \gls{f1} scores below 0.21. \ifnottechreport{\sysreweighing fails to mitigate bias while \syslfr produces a poor classifier (\gls{f1} = 0.01).}
\iftechreport{\sysreweighing fails to mitigate bias, exhibiting an \gls{eo} of 0.49. While \syslfr achieves low unfairness (\gls{eo} = 0.01), it produces a poor-performing classifier (\gls{f1} = 0.01).
\sysfairsampler performs slightly better, achieving \gls{f1} = 0.20 and \gls{eo} = 0.19, which is superior to methods not designed for \gls{le}, but still yielding models of limited utility. These results highlight the fragility of debiasing techniques when facing systematic label noise.}

Results for \gls{sb} are shown in \Cref{tab:robustness-selection-bias}. Similar to \gls{le}, \gls{sb} degrades both fairness and accuracy, though the severity of the impact varies. \sysfairshift shows the best balance between accuracy and fairness, achieving \gls{f1} = 0.35 and \gls{eo} = 0.13. \iftechreport{However, it assumes fixed marginal distributions over labels and sensitive attributes, which is often violated in real-world settings and is
mildly broken in these experiments.}
\sysreweighing increases bias (\gls{eo} = 0.36) at marginally better accuracy (\gls{f1} = 0.33). \syslfr again fails to produce a usable model (\gls{f1} = 0.01).

\begin{takeaway}
While debiasing methods like \sysfairsampler and \sysfairshift can reduce unfairness under systematic corruptions, they often fail to preserve classification performance. 
\end{takeaway}
}

\begin{table}[]\small
\begin{tabular}{l|cccc}
\toprule
Metrics         & Orig. & \sysreweighing & \syslfr & FairSampler~\cite{roh2021sample}\\ \toprule
F1 &    $0.32 \pm 0$     & $0.21 \pm 0$            & $0.01 \pm 0.02$            & $0.2 \pm 0.01$  \\
\gls{eo}  & $0.22 \pm 0$        & $0.49 \pm 0.01$            & $0.01 \pm 0.01$            & $0.06 \pm 0$ \\
\toprule
\end{tabular}
\caption{Robustness of \gls{deb} methods under \gls{le} targeting fairness measured by \gls{eo} (budget: 10\%).}\label{tab:robustness-label-bias}
\vspace{-5mm}
\end{table}

\begin{table}[]\small
\begin{tabular}{l|cccc}
\toprule
Metrics         & Orig. & \sysreweighing & \syslfr & FairShift~\cite{roh2023improving}\\ \toprule
F1 &    $0.38 \pm 0$     & $0.33 \pm 0$            & $0.01 \pm 0.02$            & $0.35 \pm 0$  \\
\gls{eo}  & $0.22 \pm 0$        & $0.36 \pm 0.01$            & $0.01 \pm 0.01$            & $0.13 \pm 0.02$ \\
\toprule
\end{tabular}
\caption{Robustness of \gls{deb} methods under \gls{sb} targeting fairness measured by \gls{eo} (budget: 10\%).}\label{tab:robustness-selection-bias}
\vspace{-5mm}
\end{table}




\begin{table}[]\small
\begin{tabular}{l|cc}
\toprule
Algorithm         & Coverage (\%) & Missing Rate (\%) \\ \toprule
\syscpsplit (clean) &    $96.2\pm 1.1$     & 0  \\
\syscpmda (clean)  & $95\pm 2.3$      & 0 \\
\syscpsplit (corrupted) &    $91.7\pm 2.1$     & $22.7\pm 6.1$  \\
\syscpmda (corrupted)  & $89\pm 2.2$ & $19.6\pm 7.9$ \\
\toprule
\end{tabular}
\caption{Actual coverage of \syscpsplit and \syscpmda with missing values in the training data ($\alpha=0.05$).}\label{tab:uqone}\vspace{-3mm}

\begin{tabular}{l|cc}
\toprule
Algorithm         & Coverage (\%) & Missing Rate (\%) \\ \toprule
\syscpsplit (clean) &    $80\pm 5.2 $    & 0  \\
\syscpmda (clean)  &    $81.6\pm 4.8$     & 0  \\
\syscpsplit (corrupted) &    $70.8\pm 4.4$     & $23.7\pm 6.4$  \\
\syscpmda (corrupted)  &    $73.5\pm 4.2$     & $20.3\pm 7.3$  \\
\toprule
\end{tabular}
\caption{Actual coverage of \syscpsplit and \syscpmda with missing values in the training data ($\alpha=0.2$).}\label{tab:uqtwo}
\vspace{-10mm}
\end{table}


\subsubsection{Uncertainty Quantification}\label{sec:benchmark:uq}
We also analyze the robustness of conformal prediction (\gls{cp}) when the training data contains \gls{mv}.
In \gls{cp} the user provides a target coverage $1-\alpha$ and the \gls{cp} approach computes a prediction interval for a test data point such that the ground-truth label is guaranteed to be within the interval with probability at least $1 - \alpha$.
We use the \dsdiabetes dataset and employ \sys using a budget of 30\% to generate \gls{mv} that break the coverage guarantee of CP. We benchmark \syscpsplit, the standard split conformal prediction, and \syscpmda \cite{zaffran2023conformal}, an extension designed for robustness to \gls{mv}.
\Cref{tab:uqone,tab:uqtwo} show the results for different target coverages $1-\alpha$. Both methods achieve the target coverage when no data errors are present.
However, with less than 30\% \gls{mv} in a single column, the average coverage of both methods dropped by more than 4.5\% when the target coverage was $0.95$ ($\alpha=0.05$), and by over 8\% for a target coverage of $0.8$ ($\alpha=0.2$).
Even though \syscpmda is designed to handle \gls{mv}, it fails to achieve the desired coverage when \emph{systematic} data errors are introduced using \sys. This is primarily due to its reliance on the assumptions that the missingness mechanism 
is conditionally independent of the label variable which is  often violated in real-world scenarios and by the \gls{mv} injected by \sys~\cite{graham2012missing}.
\iftechreport{Only under these assumptions can \syscpmda recover the marginal coverage guarantee.
However, this condition is often violated in real-world scenarios and by the \gls{mv} injected by \sys, leading to significant coverage degradation~\cite{graham2012missing}.}

\begin{takeaway}
Even \Gls{cp} approaches designed to be robust against \gls{mv} fail to maintain the coverage guarantee when \gls{mv} are systematic. 
\end{takeaway}
\subsection{Reevaluating Robustness Claims}\label{sec:exp:case-study}

\begin{table*}[t]\small
\begin{tabular}{l|cccccccc}
\toprule
Method         & Mean Imputer & Median Imputer & Iterative Imputer & KNN Imputer & H2O & Diffprep & BoostClean & AutoSklearn \\ \toprule
Random Search &    $0.8 \pm 0$     & $0.79 \pm 0$ & $0.8 \pm 0$ & $0.8 \pm 0$ & $0.61 \pm 0.15$ & $0.73 \pm 0.11$ & $0.81 \pm 0.02$ & $0.75 \pm 0.16$ \\
Beam Search  & $0.39 \pm 0.14$ & $0.36 \pm 0$ & $0.53 \pm 0$ & $0.67 \pm 0$ & $0.32 \pm 0$ & $0.37 \pm 0.02$ & $0.46 \pm 0$ & $0.4 \pm 0.16$\\
\toprule
\end{tabular}
\caption{\revm{AUC of \modellr trained on worst-case data corruptions generated by random search and beam search.}}\label{tab:sys-vs-random}
\vspace{-7mm}
\end{table*}

\begin{table}[t]
  {
    \small
    \begin{tabular}{l|cccc|c}
      \toprule
      \textbf{AUC drop}        & {[}0, 0.08) & {[}0.08, 0.16) & {[}0.16, 0.24) & $\geq 0.24$ & \textbf{Total} \\ \toprule
      \textbf{Cnt. (no rules)} & 4885                 & 514                     & 204                     & 16                     & 5619           \\
      \textbf{Cnt. (rules)}    & 1250                 & 375                     & 128                     & 16                     & 1769           \\
      \textbf{Percentage}      & 0.26                 & 0.73                    & 0.63                    & 1                      & 0.31           \\ \toprule
    \end{tabular}
  }
\vspace{-2mm}\caption{Heuristic pruning of ineffective patterns.}\label{tab:auc-drop-pct}
\vspace{-8mm}
\end{table}

Existing benchmarks have investigated the impact of systematic errors using manually specified patterns~\cite{khan2024still,schelter2021jenga}.
To evaluate whether \sys can identify vulnerabilities of \gls{ml} pipelines overlooked in prior work, we use a setting from \emph{\studyson}~\cite{khan2024still} as an example, comparing their manually specified error patterns from ~\cite{khan2024still} against patterns generated by \sys.
We use the \dsindia dataset, set the error budget to 10\%, and use  \gls{f1} as the target.
Note that \studyson assumes that both training and testing datasets contain \glspl{mv}. We conduct a search for adversarial patterns where training data errors are \gls{mnar}, while test data errors are \gls{mcar}, which is closest to our previous setting of assuming no errors in the test data.
The pattern used by \sys to identify the subpopulation to inject \gls{mv} into is:

\begin{center}\footnotesize
  \begin{minipage}{1.1\linewidth}
    \bubble{Num\_Pregnancies $\leq$ 2} $\land$ \bubble{Family\_Diabetes $=$ No} $\land$ \bubble{Type\_II\_Diabetes $=$ yes}
  \end{minipage}
\end{center}


Injecting \gls{mv} according to this pattern results in a 0.81 \gls{f1}. The \gls{f1} on the clean dataset is 0.95. However, with the same setting, the pattern tested in \studyson reports an \gls{f1} of 0.88.
This gap is larger when testing with higher error budgets.
For instance, with an error budget of 30\%, the pattern discovered by \sys leads to an \gls{f1} of 0.36, while \studyson reported an \gls{f1} of 0.87 under this setting.\iftechreport{\footnote{\studyson tested five error patterns and reported the best \gls{f1} among them.
To thoroughly identify the worst-case among these patterns for a fair comparison, we examined all five patterns. None of these lead to an \gls{f1} lower than 0.86.}}
The key difference between the patterns presented in \studyson and the one generated by \sys is that \sys also explores patterns that use the label (\text{Type\_II\_Diabetes}, in this example).
Although highly adverse, this is a realistic setting that needs to be tested.
%
\studyson uses manually created error patterns taking into account signals such as feature correlations and importance.
Although these patterns encompass ML practitioners' insights, they do not fully reflect hard real-world cases, e.g., cases where the missingness of values depends on the label.

\begin{takeaway}
Prior work is overly optimistic,  
overlooking realistic, but highly adverse, corruption types.
\end{takeaway}


\subsection{Efficiency and Effectiveness of \sys}
\label{sec:effic-effect-sys}

\revm{We evaluate the effectiveness of \sys through ablation studies (\Cref{sec:exp:ablation}) and comparison with state-of-the-art data poisoning techniques (\Cref{sec:exp:poisoning}).
We also conduct a scalability analysis and discuss the solution quality on large-scale data(\Cref{sec:exp:runtime}).}


\subsubsection{Ablation Studies}\label{sec:exp:ablation}
\iftechreportelse{We evaluate \sys's components, including beam search for \glspl{DCPT}, the TPE-based optimization for parameters, and heuristics for filtering dependencies.}{We evaluate \sys's components, including beam search for \glspl{DCPT} and the TPE-based optimization for parameters. 
We also include the ablation study for the heuristics for filtering dependencies in the extended version~\cite{techreport}.} 

\mypar{Beam Search}
We compare beam search against a baseline that
randomly samples and evaluates 100 dependency graphs, selecting the one that causes the greatest reduction in \gls{auc}. Both use TPE for the corruption parameter search.
This experiment is conducted on the \dsadult dataset using \modellr as the downstream model, with 50\% budget.
The results shown in \Cref{tab:sys-vs-random} demonstrate that \sys consistently identifies error patterns that result in significantly lower AUC compared to random search. In most cases, the margin of difference is above 0.25 in AUC. 
The huge gap arises because random sampling operates within a vast search space. 
\iftechreport{The sheer size of this search space significantly reduces the probability of identifying extremely adverse patterns through random sampling.
Furthermore, as verified in prior case studies, manually specifying error patterns based on practitioner's intuition or domain analysis is also likely to be sub-optimal with such a large search space.}


\iftechreport{
\begin{takeaway}
It is challenging to identify the adverse corruptions due to the huge search space.
Beam search effectively optimizes in this space.
\end{takeaway}
}
\reva{\mypar{TPE}
We demonstrate the effectiveness of the TPE component, which is responsible for finding adversarial corruption parameters given a dependency graph, by comparing it with the random corruption parameter search.
To do this, we randomly sample 20 dependency graphs and conduct the parameter search for each.
\Cref{fig:exp:bo-comparison} presents the AUC drop of the models trained on the corrupted data discovered by TPE and random search, where each point represents a sampled dependency graph.
\sys consistently identifies corruption parameters that lead to higher AUC drop, compared with random search, especially for the most harmful cases that \sys targets during beam search.
}
\iftechreport{
\begin{takeaway}
\reva{\sys effectively identifies corruption parameters across various scenarios.}
\end{takeaway}
}
\begin{figure}[t]
  \centering
  \begin{subfigure}{0.2\textwidth}
    \centering
    \includegraphics[width=\linewidth]{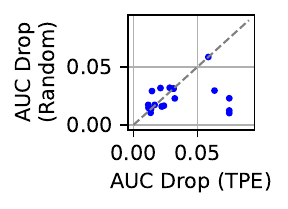}
    \vspace{-6mm}
    \caption{Logistic Regression}
    \label{fig:exp:bo-comparison:lr}
  \end{subfigure}
  \hspace{3mm}
  \begin{subfigure}{0.2\textwidth}
    \centering
    \includegraphics[width=\linewidth]{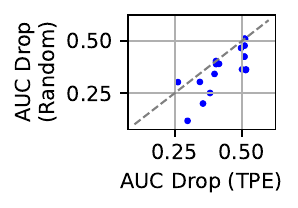}
    \vspace{-6mm}
    \caption{Decision Tree}
    \label{fig:exp:bo-comparison:dt}
  \end{subfigure}
  \begin{subfigure}{0.2\textwidth}
    \centering
    \includegraphics[width=\linewidth]{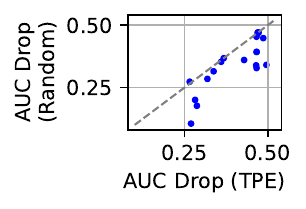}
    \vspace{-6mm}
    \caption{Random Forest}
    \label{fig:exp:bo-comparison:rf}
  \end{subfigure}
  \hspace{3mm}
  \begin{subfigure}{0.2\textwidth}
    \centering
    \includegraphics[width=\linewidth]{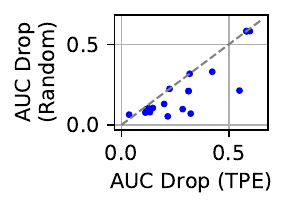}
    \vspace{-6mm}
    \caption{Neural Network}
    \label{fig:exp:bo-comparison:nn}
  \end{subfigure}
  \vspace{-3mm}
  \caption{\reva{Decrease in AUC of models trained with corrupted data, where corruption parameters are optimized by TPE (x-axis) and random search (y-axis).
  Each point represents a random dependency graph.}}\vspace{-5mm}

  \label{fig:exp:bo-comparison}
\end{figure}

\iftechreport{
\mypar{Heuristic Dependency Graph Selection}
Recall that \sys incorporates heuristic optimizations, including enforcing \gls{mnar} and always depending on the label, during beam search for efficiency.
To examine the effectiveness of these optimizations, we generate attacks based on all possible dependency graphs, injecting \gls{mv} into a single column on the \dsemployee dataset, with \syshto as the cleaning method.
For each specified dependency graph, we run \sys's \gls{bo} to obtain the corresponding worst-case corruption parameters.
As described in \Cref{tab:auc-drop-pct}, there are 5619 possible dependency graphs. After applying the heuristic pruning rules, this number drops to 1769, which reduces runtime by 68.5\%.
The dependency graphs are categorized based on their reduction in \gls{auc}.
Interestingly, the rules do not prune any of the most adverse dependency graphs. 
This result demonstrates the effectiveness of our pruning rules, and verifies the fact that (1)
\gls{mnar} is the most challenging case to deal with, and (2) errors depending on the label are more likely to be detrimental.
\begin{takeaway}
Our rules for pruning candidate dependency graphs reduce the size of the search space by a factor of $\sim 3.17$ and are effective in preserving the most adversarial dependency graphs.
\end{takeaway}
}






\begin{table}[]\small
\begin{tabular}{l|ccc}
\toprule
\% corruption         & 10 & 30 & 50\\ \toprule
\rand &    $0.85 \pm 0$     & $0.85 \pm 0$            & $0.85 \pm 0$ \\
\sysbg &    $0.83 \pm 0$     & $0.67 \pm 0$            & $0.46 \pm 0.06$ \\
\sysgc  & $0.79 \pm 0.04$        & $0.7 \pm 0.02$            & $0.45 \pm 0$ \\
\sys  & $0.58 \pm 0$        & $0.46 \pm 0$            & $0.42 \pm 0.02$ \\
\toprule
\end{tabular}
\begin{tabular}{l|ccc}
\toprule
\% corruption         & 10 & 30 & 50\\ \toprule
\rand &    $0.86 \pm 0$     & $0.86 \pm 0$            & $0.86 \pm 0$ \\
\bg &    $0.84 \pm 0.01$     & $0.78 \pm 0.01$            & $0.64 \pm 0.02$ \\
\gc  & $0.84 \pm 0.01$        & $0.81 \pm 0.01$            & $0.8 \pm 0.01$ \\
\sys  & $0.7 \pm 0.04$        & $0.34 \pm 0.12$            & $0.34 \pm 0.12$ \\
\toprule
\end{tabular}
\caption{\revm{Effect on AUC when applying \sys and indiscriminate data poisoning on \modellr (top) and \modelnn (bottom).}}\label{tab:data-poisoning}
\vspace{-10mm}
\end{table}


\begin{table*}[t]
{
  \small
\begin{tabular}{l|cccccccc}
\toprule
\textbf{Stage}    & \textbf{Mean} & \textbf{Median} & \textbf{Iterative} & \textbf{KNN} & \textbf{H2O}       & \textbf{Diffprep}  & \textbf{BoostClean} & \textbf{AutoSklearn} \\ \toprule
Dependency Search & $375.4 \pm 2.5$     & $370.2 \pm 7.7$      & $611.1 \pm 19.6$         & $651.1 \pm 6.4$   & $611.1 \pm 19.6$ & $611.1 \pm 19.6$ & $611.1 \pm 19.6$  & $611.1 \pm 19.6$   \\
Finetuning       & $26.1 \pm 0.9$      & $25.8 \pm 1$        & $30.1 \pm 3$           & $34.8 \pm 0.3$     & $12.6 \pm 0.4$    & $851.6 \pm 1$  & $153.7 \pm 2.5$   & $363.4 \pm 17$   \\
\toprule
\end{tabular}
}
\caption{Breakdown of \sys's runtime on a 30K sample of \dshmda (seconds).}\label{tab:runtime}
\vspace{-.9cm}
\end{table*}


\revm{
\subsubsection{Comparison with Data Poisoning}\label{sec:exp:poisoning}

In the following, we compare \sys with state-of-the-art data poisoning methods, as well as a random baseline (\rand), which conducts random corruption within the budget 100 times and returns the worst case. 

\mypar{Indiscriminate Attack}
Since the baselines \gc and \bg rely on editing features of data, for a fair comparison, we stick to the missing data setting, without introducing selection bias or label errors, which are more detrimental.
We leverage \sysbc for addressing missing data.
\Cref{tab:data-poisoning} presents the comparison between \sys, \bg, \gc, and \rand on the \dsadult data, where \sys consistently discovers more effective corruption than \rand, \bg and \gc.
This is primarily because existing poisoning attacks are typically designed for unstructured data without demographic attributes.
As a result, they often overlook structured patterns and rely on random sampling to select points for modification.
In contrast, \sys explicitly models systematic, non-random errors and captures their impact, particularly when corruption targets specific subpopulations.
In addition, the ineffectiveness of \rand indicates the difficulty of stochastically discovering adverse cases with completely random corruption.

Moreover, poisoning attacks are generally less effective on \modelnn than on simpler, convex models like \modellr, failing to reflect the greater sensitivity of \modelnn to biases and data quality issues. In contrast, the structured corruptions uncovered by \sys better expose this vulnerability.


\begin{takeaway}
  \revm{State-of-the-art data poisoning methods often overlook the impact of systematic subpopulation errors, thus showing worse effectiveness than \sys.}
\iftechreport{
  In addition, the vulnerability of \modelnn, revealed by \sys, is insufficiently explored by the existing data poisoning methods.
}
\end{takeaway}

\iftechreport{\mypar{Subpopulation Attack}
Recall that \subpop corrupts data via adding malicious points.
However, the range of data errors that \sys capture does not involve adding poisonous points.
For a fair comparison, we run \sys in a similar setting: introducing the same amount of label errors.
To align the goal of the task of \subpop, we use the target subpopulation AUC as the objective $\Psi$ for \sys when comparing with \subpop.
\Cref{fig:exp:subpop-vs-savage} shows the AUC drop obtained by \sys and \subpop for 20 randomly chosen subpopulations in \dsadult data, where \sys consistently finds more adverse corruption than \subpop.
In particular, in more than 90\% cases, \sys finds a corruption that leads to > 2 times AUC drop than \subpop.
The key reason for the performance gap is that \subpop always chooses the target subpopulation to corrupt.
However, the worst-case corruption that leads to low target subpopulation utility could be located in other subpopulations.
Due to the comprehensive search space of beam search, \sys successfully identifies such subpopulations other than the target subpopulation, thus showing a significant effectiveness improvement.}
}

\iftechreport{\begin{takeaway}
\revm{The corruption leading to a huge accuracy drop on the target subpopulation does not necessarily occur inside the target subpopulation.
\subpop is significantly less effective than \sys, which is primarily because \subpop only focuses on corrupting the target subpopulation}
\end{takeaway}}

\iftechreport{\begin{figure}
    \centering
    \vspace{-3mm}
    \includegraphics[width=.7\linewidth]{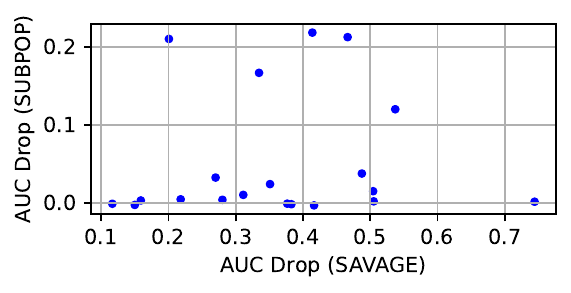}
    \vspace{-5mm}
    \caption{\revm{Decrease in subpopulation AUC due to corruption introduced by \sys (x-axis) and \subpop (y-axis).
    Each point corresponds to a randomly chosen subpopulation.}}\vspace{-5mm}
    \label{fig:exp:subpop-vs-savage}
\end{figure}}

\subsubsection{Scalability Analysis}\label{sec:exp:runtime}
We analyze the runtime breakdown of \sys, and discuss the efficiency and effectiveness of \sys when handling large-scale data with the sampling technique. \hfill\\

\begin{figure}[t]
  \centering
  \begin{subfigure}{0.22\textwidth}
    \centering
    \includegraphics[width=\linewidth]{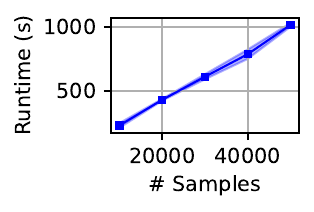}
    \vspace{-7mm}
    \caption{Dependency Search}
    \label{fig:exp:runtime:dependency}
  \end{subfigure}
  \hspace{3mm}
  \begin{subfigure}{0.2\textwidth}
    \centering
    \includegraphics[width=\linewidth]{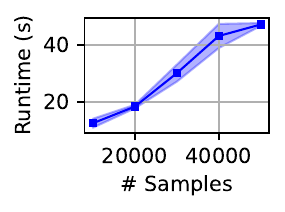}
    \vspace{-7mm}
    \caption{Finetuning}
    \label{fig:exp:runtime:finetune}
  \end{subfigure}
  \vspace{-3mm}
  \caption{\revc{Runtime of \sys with varying dataset sizes.}}\vspace{-7mm}

  \label{fig:exp:runtime-vs-size}
\end{figure}

\mypar{Runtime Breakdown}
In \Cref{tab:runtime}, we show a breakdown for \sys's runtime evaluated on a 30K sample of the \dshmda dataset with \modellr.
The runtime is broken down into two stages, where for automated cleaning techniques, including \syshto, \sysdiffprep, \sysbc, and \sysautosklearn, the dependency search involves searching a dependency graph for a proxy method instead of directly running \Cref{alg:meta-bi-level}.
The rationale for this approach is that dependency graphs typically translate well between similar cleaning techniques, and this allows us to significantly reduce the search cost by replacing an expensive cleaning technique with a cheaper proxy during search.
The finetuning stage involves tuning the corruption parameters using \Cref{alg:bo-param-search}, which is generally faster than the dependency graph search. However, as demonstrated in the comparison with \studyson, selecting the right dependency graph is critical for generating adversarial errors.

Overall, the runtime of \sys is within 20 minutes, even for expensive frameworks such as \sysdiffprep. This benefits from the utilization of proxy models during dependency search.
Without this optimization, the search time for \sysdiffprep increases to over 3 hours.
We validated the effectiveness of the proxy models by testing dependency search on \sysbc and \syshto.
Specifically, performing dependency search directly on these frameworks yielded AUC values that differed by less than 0.01 from those obtained using patterns transferred from \sysiterativeimputer. 
Another important hyperparameter, error budget, is essential for the data corruption process, but does not affect \sys's runtime much, as running the pipeline dominates the runtime of \sys.
As a result, \sys's scalability wrt. increased dataset size primarily depends on the scalability of the evaluated pipeline.
\revc{For instance, most methods, such as the \sysiterativeimputer, have linear time complexity in terms of dataset size, leading to a linear growth of \sys's runtime.
For instance, \Cref{fig:exp:runtime-vs-size} shows the runtime for \modellr and \sysiterativeimputer, and demonstrates \sys's linear complexity.}
Other than these factors, the runtime of \sys grows linearly with the beam size and number of \gls{bo} iterations, as the number of framework evaluations is linear in the number of iterations.

\begin{takeaway}
    The runtime of the evaluated frameworks dominates the runtime of \sys. By utilizing cheaper proxy models for dependency search, \sys achieves an effective and efficient search for adverse corruption mechanisms for time-consuming frameworks.
\end{takeaway}

\revc{
\mypar{Handling Large Datasets}
For large-scale datasets, the runtime of \sys is typically dominated by the cost of running the ML pipeline itself. 
According to \Cref{fig:exp:runtime-vs-size}, running \sys on the full \dshmda data is expected to take 17.8 hours.
To mitigate this, we perform the dependency search phase on a small sample of the data (1\%).
We then evaluate whether the \gls{dcp} discovered on the sample is also effective on the full dataset.
To this end, we collect all \gls{dcp}s found during the search phase of \sysiterativeimputer on the sampled \dshmda and evaluate the effectiveness of them on the full dataset.
As shown in \Cref{fig:exp:full-sample-comparison}, the model's performance on the 1\% sample closely matches that on the full dataset across all  models.
This indicates that the corruption patterns identified on the sample are also harmful at scale.
The end-to-end runtime of \sys on the sampled data is under 20 minutes, yet it discovers error patterns that cause over 0.15 AUC drop on the full dataset.}
\begin{takeaway}
    \revc{With sampling, \sys efficiently and effectively identifies adverse data corruptions for the large-scale \dshmda data that has over 3.2 million tuples.}
\end{takeaway}
\begin{figure}[t]
  \centering
  \begin{subfigure}{0.2\textwidth}
    \centering
    \includegraphics[width=\linewidth]{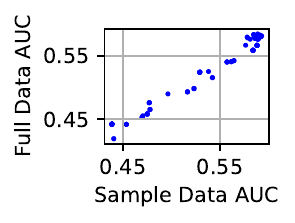}
    \vspace{-6mm}
    \caption{Logistic Regression}
    \label{fig:exp:full-sample-comparison:lr}
  \end{subfigure}
  \hspace{3mm}
  \begin{subfigure}{0.2\textwidth}
    \centering
    \includegraphics[width=\linewidth]{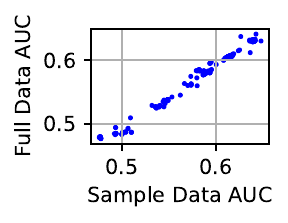}
    \vspace{-6mm}
    \caption{Decision Tree}
    \label{fig:exp:full-sample-comparison:dt}
  \end{subfigure}
  \begin{subfigure}{0.2\textwidth}
    \centering
    \includegraphics[width=\linewidth]{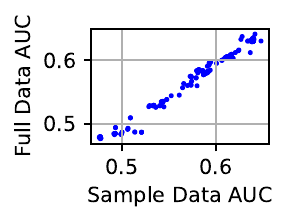}
    \vspace{-6mm}
    \caption{Random Forest}
    \label{fig:exp:full-sample-comparison:rf}
  \end{subfigure}
  \hspace{3mm}
  \begin{subfigure}{0.2\textwidth}
    \centering
    \includegraphics[width=\linewidth]{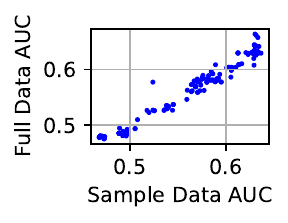}
    \vspace{-6mm}
    \caption{Neural Network}
    \label{fig:exp:full-sample-comparison:nn}
  \end{subfigure}
  \vspace{-3mm}
  \caption{\revc{AUC of models trained with different data corruption mechanisms on sample (x-axis) and full data (y-axis).}}\vspace{-5mm}

  \label{fig:exp:full-sample-comparison}
\end{figure}


\subsection{Summary}
\label{sec:exp-summary}

Our evaluation shows that all studied data cleaning, debiasing, and \gls{uq} techniques are highly sensitive to systematic errors, with even small corruptions severely degrading model performance. Certifying robustness requires a system like \sys to generate adversarial errors, as demonstrated by replicating related work. Moreover, debiasing and \gls{uq} methods often rely on assumptions that break under systematic corruption, leading to violated guarantees, even for techniques explicitly designed to handle data errors, such as \syscpmda and \sysfairsampler.
\revm{\sys achieves superior effectiveness and interpretability compared to state-of-the-art indiscriminate data poisoning techniques.}



\section{Conclusions}\label{sec:conclusions}

Data quality issues such as missing values and selection bias significantly impact ML pipelines, yet existing evaluation methods often rely on random or manually designed corruptions that fail to capture real-world systematic errors. This work introduces a formal framework for modeling the data corruption process and \sys, a system that automatically generates adversarial corruption mechanisms through bi-level optimization. \sys systematically identifies worst-case corruptions that degrade model performance while adhering to realistic constraints, providing a principled approach for evaluating the robustness of data cleaning, fairness-aware learning, and uncertainty quantification techniques. Our experiments reveal vulnerabilities in existing ML pipelines, demonstrating that current robustness measures are often insufficient against structured corruption.



\bibliographystyle{ACM-Reference-Format}
\bibliography{main}

\end{document}
\endinput